\RequirePackage{fix-cm}
\documentclass[twocolumn]{svjour3}          % twocolumn
\smartqed  % flush right qed marks, e.g. at end of proof
\usepackage{graphicx}
\usepackage{times}
\usepackage{epsfig}
\usepackage{graphicx}
\usepackage{soul}
\usepackage{enumitem}

\usepackage{amsmath, amsthm, amssymb}
\usepackage{subcaption}
\usepackage{verbatim}
\usepackage{url}
\usepackage{comment}
\usepackage{multirow}

\usepackage{balance}
\usepackage{url}
\usepackage{mathtools}
\usepackage{xspace}
\usepackage{multirow}
\usepackage{booktabs}
\usepackage{algorithm}
\usepackage{algorithmic}
\usepackage{xcolor}
\usepackage{cite}  
\usepackage{bbm}

\newcommand{\cG}{\mathcal{G}}

\newcommand{\R}{\mathbb{R}}

\newcommand{\CASE}[1]{\STATE \textbf{case} #1\textbf{:} \begin{ALC@g}}
\newcommand{\ENDCASE}{\end{ALC@g}}

\newcommand{\DEFAULT}{\STATE \textbf{default:} \begin{ALC@g}}
\newcommand{\ENDDEFAULT}{\end{ALC@g}}
\newcommand{\DEFAULTLINE}[1]{\STATE \textbf{default:} }

\DeclareMathOperator*{\argmin2}{arg\,min}

\theoremstyle{definition}
\newtheorem{exmp}{Example}

\newcommand{\matris}[1]{ \left[ \begin{smallmatrix} #1 \end{smallmatrix} \right]}
\newcommand{\Matris}[1]{ \left[ \begin{array}{cccccccccccc} #1 \end{array} \right]}

\makeatletter
\newcommand{\subalign}[1]{%
	\vcenter{%
		\Let@ \restore@math@cr \default@tag
		\baselineskip\fontdimen10 \scriptfont\tw@
		\advance\baselineskip\fontdimen12 \scriptfont\tw@
		\lineskip\thr@@\fontdimen8 \scriptfont\thr@@
		\lineskiplimit\lineskip
		\ialign{\hfil$\m@th\scriptstyle##$&$\m@th\scriptstyle{}##$\hfil\crcr
			#1\crcr
		}%
	}%
}
\makeatother
\begin{document}

\title{Implicitly Defined Layers in Neural Networks%\thanks{Grants or other notes
}


\author{Qianggong Zhang         \and
        Yanyang Gu \and
        Mateusz Michalkiewicz  \\
        Mahsa Baktashmotlagh \and
        Anders Eriksson
        %etc.
}


\institute{All Authors \at
    University of Queensland, St. Lucia, Queensland, Australia 
    \and
    Q. Zhang \at
    \email{qianggong.zhang@uq.edu.au}           
    \and
    Y. Gu \at
    \email{yanyang.gu@uq.edu.au}      
    \and
    M. Michalkiewicz  \at
    \email{m.michalkiewicz@uq.edu.au}   
   \and
    M. Baktashmotlagh \at
    \email{m.baktashmotlagh@uq.edu.au}   
    \and
    A. Eriksson \at
    \email{a.eriksson@uq.edu.au}   
}

\date{Received: date / Accepted: date}

\maketitle

%--------------------------------------------------------------------------------------

\begin{abstract}
In conventional formulations of multiple-layer feedforward neural networks, individual layers are customarily defined by explicit functions. In this paper, we demonstrate that individual layers in a neural network can be defined \emph{implicitly}. The implicitly defined layers provide much richer representation  of real problems than the standard explicitly defined ones, and consequently enable a vastly broader class of end-to-end trainable neural network architectures. We present a general framework of implicitly defined layers, and address much of the theoretical analysis of such layers with the implicit function theorem. We also show how implicitly defined layers can be seamlessly incorporated into existing machine learning libraries, particularly with respect to current automatic differentiation techniques for use in backpropagation based training. Finally, we demonstrate the versatility and relevance of our proposed approach on a number of diverse showcase problems with promising results. 
\end{abstract}

%--------------------------------------------------------------------------------------

\begin{figure}[b]
	\centering
	\begin{subfigure}[htbp]{0.2\textwidth}
		\centering
		\includegraphics[width=\textwidth]{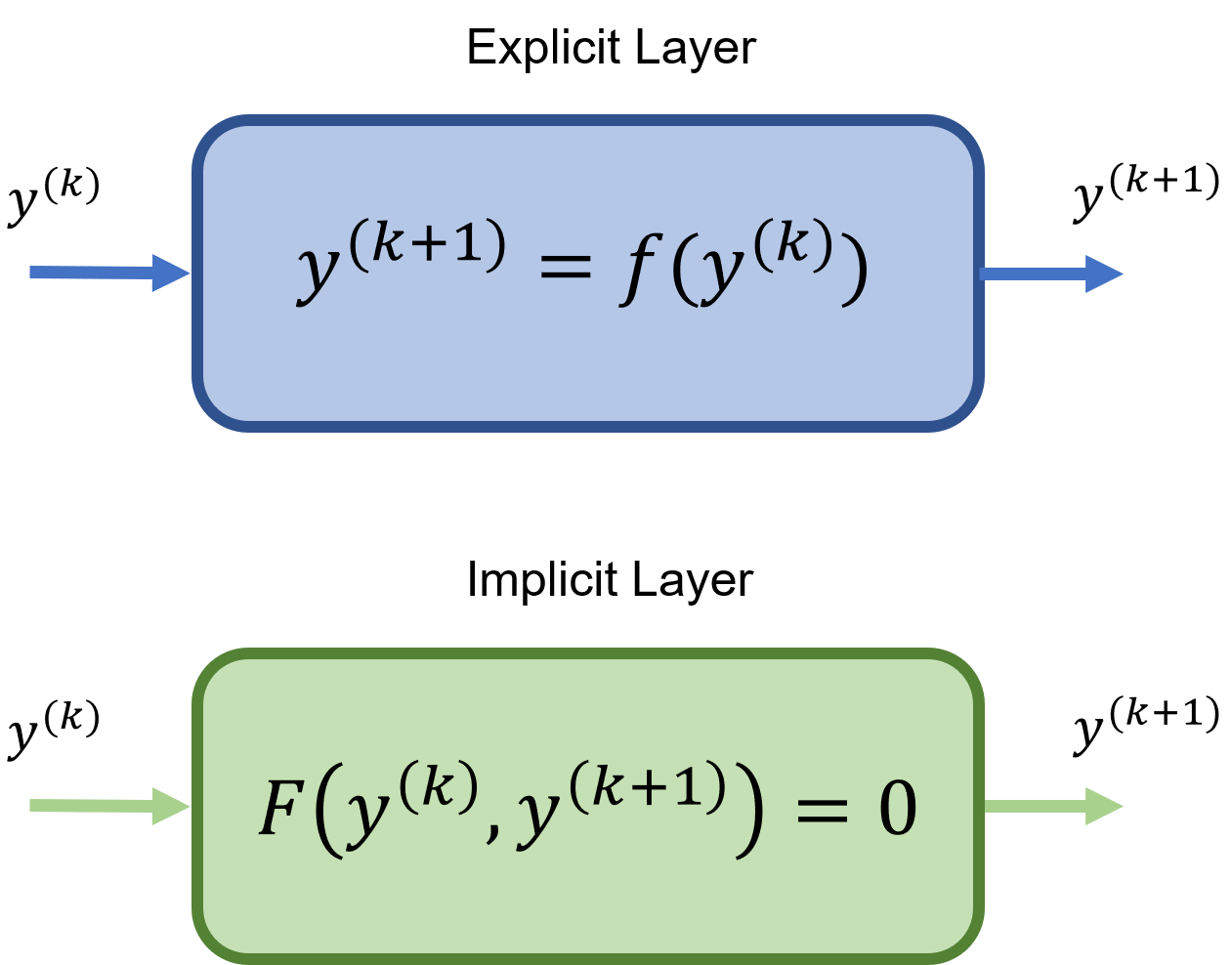}
%		\caption{}
	\end{subfigure}
	\hspace{7mm}
	\begin{subfigure}[htbp]{0.17\textwidth}
		\centering
		\includegraphics[width=\textwidth]{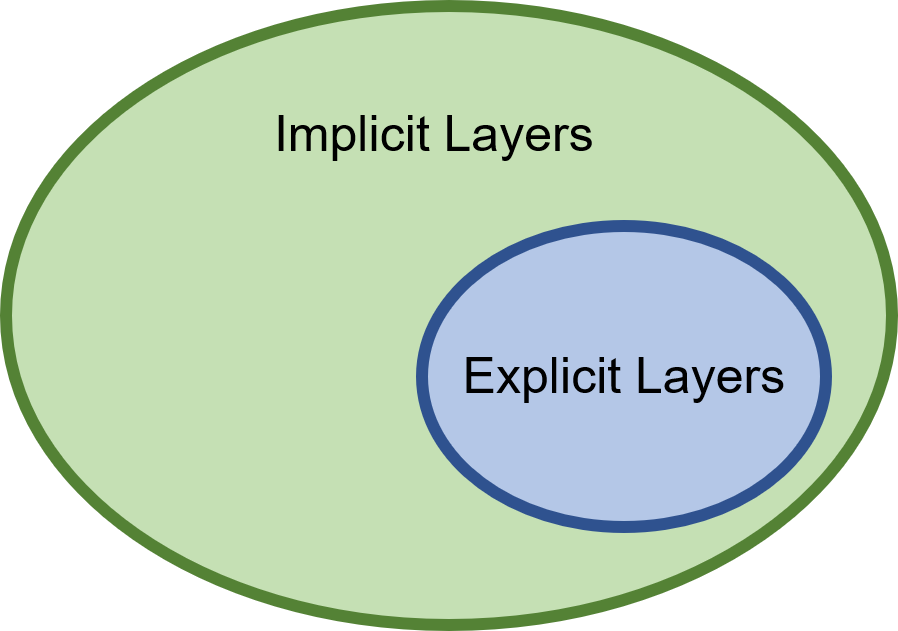}
%		\caption{}
	\end{subfigure}
	\caption{Explicitly vs Implicitly defined layers (left). 
	The latter enables a broader class of end-to-end trainable networks (right). }
	\label{fig:intro_fig}
\end{figure}

\section{Introduction}
\label{intro}

Conventional multiple-layer neural networks are entirely defined through \emph{explicit} expressions of its entering layers and loss functions. These expressions are typically provided in the form of a function that maps the input $y^{(k)}$ 
of the $k^{th}$ layer to its output $y^{(k+1)}$, as 
\begin{equation}
y^{(k+1)} = f(y^{(k)}). 
\end{equation}
Here $y^{(k)}$ may contain the output of the previous layer as well as the trainable parameters (commonly denoted $\theta$).  

This explicit approach has the advantage that training through back-propagation, a method that operates on the partial derivatives associated with each layer, is straightforward to implement. However, this approach has also proven to be rather restrictive in a sense that 
there are limited types of the layers that can be included in an end-to-end trainable network. 

This work investigates the implicit approach.  The term \emph{implicit layer} refers to a neural network layer that is defined implicitly by an implicit equation: %, that is 
\begin{equation}
F(y^{(k)},y^{(k+1)}) = 0. 
\end{equation}
An intuition into why this reformulation is beneficial can perhaps be found in multi-variable calculus, where the notion of \emph{functions given by a formula} has invariably been seen as too limited for many purposes. There are countless examples of functions that cannot be expressed explicitly, for instance, the locus of the expression 
\begin{equation}
y^5 + 16y- 32x^3 + 32x = 0
\end{equation}
defines a precise and sketchable subset of $\R^2$, yet no formula for it exists. 

As a matter of fact, the set of implicit functions is a proper superset of the set of explicit functions.  This follows trivially as any explicit function in the form $y^{k+1}=f(y^k)$ can be defined implicitly as $y^{k+1}-f(y^k)=0$.  As feedforward layers are conceptually functions mapping the input onto the output, a similar conclusion can also be made here. That is, not all implicit layers can be expressed explicitly. The reverse, however, is indeed true, that all explicit layers can be expressed implicitly. Figure \ref{fig:intro_fig} illustrates the premise.

%\subsection{Contribution}
%\vspace{1mm}
%\noindent

This work presents a general framework of implicitly defined layers.  Section~\ref{sec:implicit} addresses much of the theoretical analysis of implicit layers through the implicit function theorem.  Section~\ref{sec:prop_through_imp_layers} describes the treatment of backpropagation of implicit layers; specifically, it is demonstrated how our framework is directly applicable to current automatic differentiation techniques for use in backpropagation based training.  In section~\ref{sec:experiments} a number of diverse showcases demonstrate the versatility and practical benefit of the proposed approach.

%--------------------------------------------------------------------------------------

\section{Related Works}
\label{sec:review}

Optimization plays a key role in a wide array of machine learning applications as a tool to perform inference in learning. Differentiation through optimization problems, e.g., $\argmin2$ 
operators, has seen a number of advances in recent years, among which, there are techniques that come up in bi-level optimization~\cite{gould2016differentiating,kunisch2013bilevel} and sensitivity analysis~\cite{bonnans2013perturbation,johnson2016composing,mairal2011task}.

More specifically, ~\cite{kunisch2013bilevel} proposed semi-smooth Newton algorithms that could efficiently find optimal regularization parameters, leading to efficient learning algorithms. The proposed bi-level learning framework could be applied to variational models, including the non-smooth functions, but not including data fidelity terms that are different from quadratic ones. The authors of ~\cite{gould2016differentiating} presented results for differentiating parameterised $\arg\!\min$ and $\arg\!\max$ optimization problems through equality constraints, but did not consider inequality constraints, and thus could only be applied to a limited class of problems with smooth functions within the $\arg\!\min$ and $\arg\!\max$ domain. The work of~\cite{mairal2011task} considered $\arg\!\min$ differentiation for a dictionary learning problem, and presents an efficient algorithm to solve it.

Moreover, \cite{amos2017input} considered $\arg\!\min$ differentiation within the context of the bundle method, and learned the inference step along with the network itself, without building structured prediction architectures explicitly. ~\cite{johnson2016composing} used implicit differentiation on convex objectives with coordinate subspace constraints but was unable to cope with general linear equality constraints and inequality constraints.

All the aforementioned approaches have limited applications in a sense that they either consider equality constraints in their implicit differentiation~\cite{gould2016differentiating,kunisch2013bilevel} rather than both equality and inequality constraints, or they can only insert the optimisation problem in the final layer of the network~\cite{johnson2016composing}.

Most closely related to our work is the recent method of~\cite{amos2017optnet}, in which, the implicit differentiation can be performed through both inequality and equality constraints, and the optimisation problems can be inserted anywhere in the network. To derive the gradients from the KKT matrix of the optimisation problem, OptNet~\cite{amos2017optnet} makes use of techniques from matrix differential calculus. However, this work is restricted to convex quadratic problems only.

This work differs from the existing works in that we are proposing a more general framework applicable to any layer expressible as an implicit function. Furthermore, most of the above mentioned work requires manual derivations and implementation of analytic gradient expressions, which is not needed in our framework.

%--------------------------------------------------------------------------------------------

\section{The Implicit Layer}
\label{sec:implicit}
%\section{Implicit Functions and Implicit Layers}

In this work, we present a principled treatment of \emph{implicitly} defined layers in 
feedforward neural networks. We formally define this concept as follows. 
\begin{definition}[Implicit Layer]
A neural network layer is implicitly defined if its output $y^{(k+1)}\in\R^m$ is given 
as the unique solution of the system of equations $F: \R^n \times \R^m \mapsto \R^m$,
	\begin{align}
		F(y^{(k)},y^{(k+1)})=0,
		\label{implicit_def}
	\end{align}
for some input  $y^{(k)}\in\R^n$. 
\end{definition}
\noindent
We distinguish this from the usual explicitly defined feedforward layers 
where the relationship between input and output is given as $y^{(k+1)}=f(y^{(k)})$. 
As before,  $y^{(k)}$ does not only denote the output of the previous layer but also 
the trainable parameters of the current layer.

\subsection{The Implicit Function Theorem}

To overcome the limitations of the naive definition of functions as explicit expressions, functions are instead commonly defined in a set-theoretic sense \cite{hamilton_1983}. 

\begin{definition}[Function]\label{def:fun}
Here a function $f$ from a set $X$ to a set $Y$ is formally defined as a set of ordered pairs 
$(x,y)$, $x\in X$ and $y\in Y$ with the properties that 
\begin{itemize}
    \item for each $x\in X$ there exist a pair $(x,y)\in f$;
    \item if both $(x,y_1)\in f$ and $(x,y_2)\in f$, then $y_1=y_2$.
\end{itemize}
\end{definition}
\noindent
With this definition, each $x\in X$ defines a unique $y\in Y$ for which $(x,y)\in f$. 
That is, it describes the process of associating each element of $X$ with a single 
unique element in $Y$. It is common to use the more convenient notation of letting $y=f(x)$ 
denote $(x,y)\in f$. 

%Many of the technical issues can be resolved by instead adopting a set-theoretic approach, \cite{}. 
%Here a function is defined in terms of the graph of the function rather than the naive formulaic 
%definition. This set-theoretic approach has the advantage of providing a generalized and unifying 
%framework capable of handling explicit as well as implicit representations of functions equally. 
%For the remainder of this paper we will use this definition of functions. 

The system of equations in \eqref{implicit_def} define an arbitrary closed (if $F$ is continuous) subset of $\R^m$. Although no explicit expression might exist, it can be shown that, under certain conditions, such implicit expressions can be \emph{locally} expressed as functions (with Definition~\ref{def:fun}).  The details of sufficient conditions for this to hold is provided by the \emph{Implicit Function Theorem}~\cite{krantz2012implicit}, see theorem~\ref{implicit_function_theorem}. 

\begin{theorem}[Implicit Function Theorem]
\label{implicit_function_theorem}
Given three open sets $X \subseteq \R^n, Y \subseteq \R^m$, and $Z \subseteq \R^m $, if function $F : X \times Y \mapsto Z$ is continuously differentiable, and $(\hat{x}, \hat{y}) \in \R^{n} \times \R^m$ is a point for which
\begin{align}
F(\hat{x}, \hat{y}) = \hat{z},
\end{align}
and the Jacobian of $F$ with respect to $y \subseteq Y$
\begin{align}
J_{F, y}\Big|_{i,j}=\left[{\frac {\partial F_{i}}{\partial y_{j}}}\right]
\end{align}
is invertible at $(\hat{x}, \hat{y})$, then there exists an open set $W \subset \R^{n}$ with $x \in W$ and a unique continuously differentiable function $\phi : W \mapsto Y$ such that $y = \phi(x)$ and
\begin{align}
F(x, y) = \hat{z} 
\end{align}
holds for $x \in W$.
\end{theorem}

In addition, it can be shown that the partial derivatives of $\phi$ in $W$ are given by 
%\begin{align}
%{\frac {\partial y}{\partial x_l}}=-\left[J_{F,y}\right]^{-1}\left[{\frac {\partial F}{\partial x_{l}}}\right],
%\end{align}
%which leads to the compact form
\begin{align}\label{eq:thm1_J}
J_{y, x}=-\left[J_{F,y}\right]^{-1}\left[J_{F,x}\right].
\end{align}

This theorem states that under certain mild conditions on the partial derivatives, the solution to a system such as \eqref{implicit_def} is locally the graph of a function. Note that these functions might also only be available implicitly. However, according to this theorem, if such functions exist, they must be continuously differentiable and their derivatives can have a simple analytical expression~\eqref{eq:thm1_J}. Most of the results in this paper will build on this latter consequence of the implicit function theorem.

%\subsubsection{Invertibility of the Jacobian} 
%\qg{Could do.............}
%1. A diagram with indifferentiable examples.\\
%2. Connect indifferentiability to non-invertible Jacobian (From the implicit function theorem).\\
%3. Solution: subgradient and pseudo-inverse.

%Next we will show how an implicitly defined layer can be realised and incorporated into a trainable feedforward network, mainly through the application of the above theorem. 

%\subsection{Forward and Backward Passes in Implicit Layers}
%\section{Backpropagation in Implicit Layers}
\section{Propagating through Implicit Layers}
\label{sec:prop_through_imp_layers}
\subsection{The Forward Pass}
As in conventional neural networks pipelines, in our proposed approach, the forward path and the backward path 
of an implicit layer are independent.  
The forward pass in an implicit layer is directly realised through the solution of \eqref{implicit_def}. 
The most appropriate choice of solver is highly task specific, hence we will assume that a method of performing the forward pass is given along with the an implicit definition of a layer. 
Our proposed framework is entirely agnostic to the choice of forward pass solvers, we can therefore make this
assumption without loss of generality. 
Examples of different forward pass solvers are given in section \ref{sec:experiments}.

\subsection{The Backward Pass}
\label{sec:backprop}
The remaining question then relates to the backward pass through the implicit layer. To form a backward pass of a neural network layer we require the partial derivatives of its output with respect to its input, including the previous layer's output and all the trainable parameters of this layer. Hence, we need an expression for all these partial derivatives and an efficient way to calculate them. We will show that the former is provided by the implicit function theorem and the latter can be obtained by utilising existing automatic differentiation techniques. 

%\noindent
%\textbf{The Backward Pass}
The backward pass of an implicit layer is obtained as follows.  Let the current state of the layer be given by $(\hat{y}^{(k)}, \hat{y}^{(k+1)})$ such that $F(\hat{y}^{(k)}, \hat{y}^{(k+1)})=c$, where $c\subseteq \R^n$ a vector of constants.  Our premise is that there then exists, in the set-theoretic sense, a function $\phi: \R^m \mapsto \R^n$ such that $y^{(k+1)}=\phi(y^{(k)})$ and that $\phi$ is differentiable in some neighbourhood of $(\hat{y}^{(k)}, \hat{y}^{(k+1)})$. 

Let the partial Jacobian of $F$ with respect to the output $y^{(k+1)}$ be denoted by 
\begin{align}
%J_{F, y^{(k+1)}}=\Matris{\frac{\partial F_1}{\partial y_1^{(k+1)}} & \dots & \frac{\partial F_1}{\partial y_m^{(k+1)}} \\ \vdots & \ddots & \vdots %\\\frac{\partial F_m}{\partial y_1^{(k+1)}} & \dots & \frac{\partial F_m}{\partial y_m^{(k+1)}} }.
J_{F, y^{(k+1)}}\Big|_{i,j}=\left[{\frac {\partial F_{i}}{\partial y^{(k+1)}_{j}}}\right],
\end{align}
then from the the implicit function theorem, theorem~\ref{implicit_function_theorem}, and the 
differentiability assumption on $\phi$, 
$\left[J_{F,y^{(k+1)}}\right]$ will have full rank at $(\hat{y}^{(k)}, \hat{y}^{(k+1)})$ 
and the sought partial derivatives of $\phi$ are given by
% it follows directly that if $\left[J_{F,y^{(k+1)}}\right]$ has full rank at $(\hat{y}^{(k)}, \hat{y}^{(k+1)})$ then our assertion will hold, that a differentiable $\phi$ exists and is given by
%
%\begin{align}\label{eq:backward_pass}
%\begin{aligned}
%&\frac{\partial y^{(k+1)}}{\partial y_j^{(k)} }(y^{(k)}_0, y^{(k+1)}_0) \\
%=~& -\left[ J_{F,y^{(k+1)}}(y^{(k)}_0, y^{(k+1)}_0)\right]^{-1}\left[\frac{\partial F}{\partial y_j^{(k)}}(y^{(k)}_0, y^{(k+1)}_0)\right]
%\end{aligned}
%\end{align}
%
%for all $j=1,...,m$, which leads to the compact form
\begin{align}\label{backward_pass}
\begin{aligned}
J_{y^{(k+1)},y^{(k)}} =~& -\left[ J_{F,y^{(k+1)}}\right]^{-1}\left[ J_{F,y^{(k)}}\right],
\end{aligned}
\end{align}
evaluated in some neighbourhood of $(\hat{y}^{(k)}, \hat{y}^{(k+1)})$.

The Jacobian of the output with respect to the input is the key to applying the chain rule to compute the derivatives of the loss $L$ with respect to the input $\hat{y}^{(k)}$ and so to propagate gradients backwards:
\begin{align}\label{eq:chain_rule}
\begin{aligned}
\left(\frac{\partial{L}}{\partial{\hat{y}^{(k)}}}\right)^T &= \left(\frac{\partial{L}}{\partial{\hat{y}^{(k+1)}}}\right)^T \left[J_{y^{(k+1)},y^{(k)}}\right]\\
&= - \left(\frac{\partial{L}}{\partial{\hat{y}^{(k+1)}}}\right)^T \left[ J_{F,y^{(k+1)}}\right]^{-1}\left[ J_{F,y^{(k)}}\right].
\end{aligned}
\end{align}
It is the above expression that underpins our formal treatment of implicitly defined layers and is what
permits their inclusion into standard backpropagation training techniques.

%\subsubsection{Conformity to the Explicit Case} 
%\subsubsection{Relationships to Explicitly Defined Layers} 

As previously discussed any explicit layer can also be defined implicitly. 
This then implies that the resulting backward pass for such a layer should be 
invariant to the manner in which it is defined, i.e. explicitly or implicitly. 
To verify that Equation~\eqref{eq:chain_rule} conforms to the standard treatment for explicit layers, consider an explicit layer that is defined by function $y^{(k+1)} = f(y^{(k)})$ and is currently at the state $(\hat{y}^{(k)}, \hat{y}^{(k+1)})$.  The 
implicit form of this layer is thus
\begin{align}\label{eq:explicit_fun}
F(y^{(k)}, y^{(k+1)}) = f(y^{(k)}) - y^{(k+1)} = 0.
\end{align}
It follows from~\eqref{eq:explicit_fun} that 
\begin{align}
\begin{aligned}
    \left[ J_{F,y^{(k+1)}}\right] = -I~~\text{and}~ \left[ J_{F,y^{(k)}}\right] = \left[ J_{f,y^{(k)}}\right],
\end{aligned}
\end{align}
which, substituted into~\eqref{eq:chain_rule}, leads to the familiar equation for explicit layer backpropagation
\begin{align}\label{eq:explicit_chain}
\begin{aligned}
\left(\frac{\partial{L}}{\partial{\hat{y}^{(k)}}}\right)^T = \left(\frac{\partial{L}}{\partial{\hat{y}^{(k+1)}}}\right)^T \left[ J_{f,y^{(k)}}\right].
\end{aligned}
\end{align}

Lastly, as stated in \eqref{eq:chain_rule},  calculating $\left[J_{y^{(k+1)},y^{(k)}}\right]$ requires the explicit construction of 
the Jacobian $\left[J_{F,y^{(k)}}\right]$. As this matrix is typically 
large this operation can be very costly. 
Modern deep learning packages seldom, if at all, explicitly construct 
%the Jacobian $\left[J_{f,y^{(k)}}\right]$ in~\eqref{eq:explicit_chain}, 
the entering matrices in the backward pass, but instead derive 
Vector-Jacobian Products for the right hand side of~\eqref{eq:explicit_chain}.
This technique results in greatly reduced memory requirements and computational 
costs. 
%Same techniques are applicable to backpropagating implicit layers too.  
This approach is applicable to backpropagating implicit layers as well.  
Note that the right hand side of~\eqref{eq:chain_rule} is also a Vector-Jacobian Product of the vector 
$-\left(\partial{L}/\partial{\hat{y}^{(k+1)}}\right)^T \left[ J_{F,y^{(k+1)}}\right]^{-1}$ 
and Jacobian matrix $\left[ J_{F,y^{(k)}}\right]$, thus it is feasible to improve  computational efficiency by carefully analysing the expression of and the structure of  $\left[J_{F,y^{(k)}}\right]$; see Section~\ref{sec:showcase_mnist} for an example of efficiently backpropagating an implicit layer doing quadratic programming.

\subsection{Automatic Differentiation}\label{sec:auto_diff}

Deriving the analytical expression for the implicit backward pass 
%$\left[J_{F,y^{(k+1)}}\right]$ 
can be prohibitively time consuming and error-prone, particularly in situations where the models or network architecture is expected to change frequently.  On the other hand, techniques for \emph{automatic differentiation} in deep learning packages provide accurate, efficient, and reliable computation of partial derivatives in a fully automated manner, thus eliminate the need for manual derivation and implementation of analytical gradient formulae.  In light of this, we propose a method by which the partial gradients of implicit layers could be automatically calculated.

In this section we show how existing implementations of automatic differentiation can be used to provide backward passes through implicitly defined layers with little modification.  

Let our implicit layer be defined as in \eqref{implicit_def}. Now consider the related explicit layer defined by 
\begin{align}\label{eq:autodiff0}
\begin{aligned}
&z^{(k+1)} = F(z^{(k)}), ~~\text{where}\\
&z^{(k)} = [y^{(k)},y^{(k+1)}]\in \R^{n+m}~~\text{and}~~z^{(k+1)}\in 0^m.
\end{aligned}
\end{align}
As this layer is defined explicitly we can apply existing automatic differentiation methods directly to yield the partial derivatives 
%This will give us 
\begin{align}
\frac{\partial z_i^{(k+1)}}{\partial z_j^{(k)}}= \frac{\partial F_i}{\partial z_j^{(k)} },  \hspace{4mm} i\in[1,m],\ j\in[1,m+n], 
\label{eq:autodiff1}
\end{align}
or more compactly
\begin{align}
\frac{\partial z^{(k+1)}}{\partial z^{(k)}}= 
\Matris{ 
J_{F, y^{(k+1)}} %(y^{(k)},y^{(k+1)})  
\ \Big| \
J_{F, y^{(k)}} %(y^{(k)},y^{(k+1)})  
}.
\label{eq:autodiff2}
\end{align}

%Here the vertical bar between two Jacobians represents the operation of matrix concatenation. 

Comparing \eqref{eq:autodiff2} with \eqref{backward_pass}
we note that the elements $J_{F, y^{(k)}}$ and $J_{F, y^{(k+1)}}$ for calculating the backward pass in \eqref{backward_pass}
are provided by automatic differentiation of an explicit layer as of \eqref{eq:autodiff1}. 
Consequently, the automatic differentiation of implicit layers
can be realised as a matrix inversion and multiplication, with no need 
for manual derivation or model specific implementations. 

%Let us consider the following simple example of an implicitly defined layer. 
\noindent
Let us consider the following simple example: 
%\begin{proof}[Example 1]
\begin{exmp}
%In this section we provide a simple example of We consider an implicit layer defined by the follow 
%simple example of a system of implicit functions $F(x,y) = 0$ 
\begin{align}
\begin{aligned}
	&F_1(x,y)=x^2 + y_1^2 + y_2^2 - 4 , \\
	&F_2(x,y)= xy_1 - 1.
\end{aligned}
\end{align}
A solution to $F(x,y)=0$ is given by $(\hat{x}=1, ~\hat{y}=[1, \sqrt{2}])$.
To calculate the backward pass, (i.e. the partial derivatives of the output $y$ with respect to the input $x$) through such a layer we instead look at the related explicit layer defined by 
\begin{align}
z^{(k+1)} = F(z^{(k)}),~~~ z^{(k)}\in \R^3, \ z^{(k+1)}\in \R^2,
\end{align}
and specifically in this example
\begin{align}
\begin{aligned}
&z^{(k)} = (x, y) = (\hat{x}, \hat{y}_1, \hat{y}_2), \\
&z^{(k+1)} = (0,0).
\end{aligned}
\end{align}
As the layer is now defined explicitly, we can apply automatic differentiation directly to provide 
\begin{align}
\frac{\partial z_i^{(k+1)}}{\partial z_j^{(k)}}(z)= \frac{\partial F_i}{\partial z_j^{(k)}}(z),  \hspace{4mm} i=1,2,\ j=1,2,3, 
\label{eq:autodiff_1}
\end{align}
for some $z\in \R^3$.
In compact form, this equation reads
\begin{align}
\left[J_{z^{(k+1)}, z^{(k)}}(z)\right] = \left[J_{F, z^{(k)}}(z)\right].
\end{align}
The term on the right hand side of the above equation is instantiated as
% \begin{align}\label{eq:dev}
% \begin{aligned}
% \left[J_{F, z^{(k)}} (z^{(k)})\right]
% &= 
% \left[ \arraycolsep=1.8pt\def\arraystretch{2.2} 
% \begin{array}{cccc}
% 	\frac{\partial F_1}{\partial x} & \frac{\partial F_1}{\partial y_1} & \frac{\partial F_1}{\partial y_2} \\
% 	\frac{\partial F_2}{\partial x} & \frac{\partial F_2}{\partial y_1} & \frac{\partial F_2}{\partial y_2}
% \end{array} \right]
% %= \Matris{ J_{F, x}(z^{(k)})~ \Big| ~J_{F, y}(z^{(k)})}\\
% = [ J_{F, x}(z^{(k)})~ \Big| ~J_{F, y}(z^{(k)})]\\
% &=
% \left[ \arraycolsep=1.8pt\def\arraystretch{2.2} 
% \begin{array}{cccc}
% 	2x & 2y_1 & 2y_2 \\
% 	y_1 & x & 0
% \end{array} \right]
% = 
% \left[ \arraycolsep=1.8pt\def\arraystretch{2.2} 
% \begin{array}{cccc}
% 	2 & 2 &2\sqrt{2}\\ 
% 	1 & 1 & 0
% \end{array} \right].
% \end{aligned}
% \end{align}
\begin{align}\label{eq:dev}
\begin{aligned}
\matris{
J_{F, z^{(k)}} (z^{(k)})
 }
&= 
\left[ \arraycolsep=1.8pt\def\arraystretch{2.2} 
\begin{array}{cccc}
\frac{\partial F_1}{\partial x} & \frac{\partial F_1}{\partial y_1} & \frac{\partial F_1}{\partial y_2} \\
	\frac{\partial F_2}{\partial x} & \frac{\partial F_2}{\partial y_1} & \frac{\partial F_2}{\partial y_2}
\end{array} \right]
%= \Matris{ J_{F, x}(z^{(k)})~ \Big| ~J_{F, y}(z^{(k)})}\\
%= [ J_{F, x}(z^{(k)})~ \Big| ~J_{F, y}(z^{(k)})]\\
= \matris{ J_{F, x}(z^{(k)})~ & \Big| & ~J_{F, y}(z^{(k)})}\\
&=
\left[ \arraycolsep=1.8pt\def\arraystretch{1.8} 
\begin{array}{cccc}
2x & 2y_1 & 2y_2 \\
	y_1 & x & 0
\end{array} \right]
= 
\left[ \arraycolsep=1.8pt\def\arraystretch{1.8} 
\begin{array}{cccc}
	2 & 2 &2\sqrt{2}\\ 
	1 & 1 & 0
\end{array} \right].
\end{aligned}
\end{align}
And finally, the partial derivative of the output $y$ of the implicit layer with respect to the input $x$ is calculated by~\eqref{backward_pass}:
\begin{align}
\begin{aligned}
\left[J_{y,x}([\hat{x},\hat{y}])\right] &= -\big[J_{F,y}(\hat{x},\hat{y})\big]^{-1} \big[J_{F,x}(\hat{x},\hat{y})\big] \\
&=-\left[ \arraycolsep=1.8pt\def\arraystretch{2.2} 
\begin{array}{cc}
2 &2\sqrt{2}\\ 
1 & 0
\end{array} \right]^{-1}
\left[ \arraycolsep=1.8pt\def\arraystretch{2.2} 
\begin{array}{c}
2\\ 
1
\end{array} \right]
\end{aligned}
\end{align}

Recall that as ~\eqref{eq:dev} is supported by the automatic differentiation implementations of over-the-shelf machine learning libraries, we eventually get the partial derivative $\left[J_{y,x}\right]$ without manually deriving algebraic expressions of the derivatives.
%\qedsymbol
%\qedhere
\pushQED{\qed} 
\qedhere
\popQED
\end{exmp}

\section{Example Applications}\label{sec:experiments}
%\section{Application Showcases}\label{sec:experiments}
%\section{Showcases}\label{sec:experiments}
In this section we present a number of example applications to demonstrate the usage and features of the proposed framework for implicit layers.  
In particular, we show: 
how to model a real problem as implicit functions;
how implicit layers enables end-to-end training;
how the backward path is independent to the forward solver; and
how the automatic differentiation feature copes with complex functions.

Note, our intention here is not to propose new methods for solving specific tasks nor attempting to improve  state-of-the-art algorithms but rather to highlight the versatility and accessibility of the proposed framework.\footnote{Demo codes are available on \url{github.com/qgzhang/Imp_layers_demo}.}

% \section{Example Applications}\label{sec:experiments}
% %\section{Application Showcases}\label{sec:experiments}
% %\section{Showcases}\label{sec:experiments}
% In this section we present a number of example applications to demonstrate the usage and features of the proposed framework for implicit layers.  In particular, we show 
% \begin{itemize}
%     \item how to model a real problem as implicit functions,
%     \item how implicit layers enables end-to-end training,
%     \item how the backward path is independent to the forward solver, and
%     \item how the automatic differentiation feature copes with complex functions.
% \end{itemize}
% Our intention here is not to propose new methods for solving specific tasks nor attempting to beat state-of-the-art algorithms but rather to highlight the versatility and accessibility of the proposed framework.

\subsection{Quadratic Programming Layers }
\label{sec:showcase_mnist}

We begin with an introductory example showing how the proposed framework 
models a Quadratic Programming (QP) problem as an implicit layer and 
then determines the necessary expressions for backpropagation
for use in an end-to-end trainable neural network.
Including QP in such a manner was also proposed 
in the notable work~\cite{amos2017optnet} as a way to encode 
constraints and dependencies that conventional explicit layers are unable to capture.
The efficiency of this approach was demonstrated on a number of problems 
including, signal denoising and handwritten digit recognition. However, this work
is restricted to convex quadratic problems only, in addition it also requires the 
backward step to be explicitly implemented. These are limitations our proposed 
framework do not possess. 

To better illustrate our example, we follow the existing work of 
OptNet~\cite{amos2017optnet} closely but elaborate relevant 
concepts from the point of view of implicit layers.  
We define a convex QP as 
\begin{align}\label{eq:QP}
	\begin{aligned}
		& \underset{y}{\arg\!\min}~~\frac{1}{2}y^TQy + q^Ty\\
		& s.t.~ Ay = b,~ Gy <= h,
	\end{aligned}
\end{align}
where $y$ is the optimisation variable; and $Q\succeq 0, q, A, b, G$, and $h$ are parameters 
of the QP problem.  

$Q, q, A, b, G$, and $h$ collectively represent both the input and the trainable 
parameters of the layer. They may or may not dependent on the input of the layer 
but, conceptually, they can all be classified as explicitly differentiable functions 
of the input. Therefore, the key to backpropagate this layer is to determine the derivatives of the loss $L$ with respect to $Q, q, A, b, G$, and $h$. 
The remaining task is then simply direct application of the standard chain rule.

Using the KKT conditions we can write~\eqref{eq:QP} into a system of implicit 
functions. 
The Lagrangian function of~\eqref{eq:QP} is given by
\begin{align}
	\begin{aligned}
		L(y, \lambda, \nu) = \frac{1}{2}y^TQy + q^Ty + \lambda^T(Ay-b) + \nu^T(Gy-h),
	\end{aligned}
\end{align}
with dual variables $\lambda$ and $\nu$.
The corresponding KKT conditions then become
%implicit functions $F$ are defined by the KKT equations
\begin{align}\label{eq:QP_KKT}
F:
\left\{
	\begin{aligned}
		 Qy + q + A^T\lambda +  G^T\nu&=0,   &(stationarity)\\
		 Ay-b&=0,   &(feasibility) \\
		 D(\nu)(Gy-h)&=0,    &(compl.\ slackness)
	\end{aligned}
	\right.
\end{align}
where $D(\nu)$ denotes a matrix with $\nu$ in the diagonal and $0$ everywhere else.  
As \eqref{eq:QP_KKT} forms the necessary conditions for the solution of 
\eqref{eq:QP} it also forms a system of equations that we can be used to define
an implicit representation of the QP layer, i.e., $F$ as in \eqref{implicit_def}. 
Note that under the above definition, the primal $y$ and the dual $\lambda$ and 
$\nu$ correspond to the output of the layer, i.e., $y^{(k+1)}$ as in~\eqref{eq:chain_rule}; and $Q, q, A, b, G$ and $h$ correspond to the input of the layer, i.e., $y^{(k)}$ 
in~\eqref{eq:chain_rule}:
\begin{align}
	\begin{aligned}
		&y^{(k+1)} = (y, \lambda, \nu)\\
		&y^{(k)} = (Q, q, A, b, G, h).
	\end{aligned}
\end{align}
Now, in order to apply~\eqref{eq:chain_rule} to complete the backpropagation process, we need $J_{F, y^{(k)}}$ and $J_{F, y^{(k+1)}}$, which are obtainable by differentiating~\eqref{eq:QP_KKT}:
\begin{align}\label{eq:JFyk1}
	\begin{aligned}
		\left[J_{F, y^{(k+1)}}\right] &= \left[ J_{F, y} \Big| J_{F, \lambda} \Big| J_{F, \nu} \right]\\
		&= \left[	
    		    \begin{array}{c}
                    Q \\
                    A \\   
                    D(\nu)G
                \end{array}
                \Bigg|
                \begin{array}{c}
                    A^T \\
                    0 \\   
                    0
                \end{array}
                \Bigg|
                \begin{array}{c}
                    G^T \\
                    0 \\   
                    D(Gy-h)
                \end{array}
		\right]
	\end{aligned}
\end{align}		
and
\begin{align}\label{eq:JFyk}
	\begin{aligned}		
		\left[J_{F, y^{(k)}}\right] &= \left[ J_{F, Q} \Big| J_{F, q} \Big| J_{F, A} \Big| J_{F, b} \Big| J_{F, G} \Big| J_{F, h}\right]\\ 
		&= \left[	
                \begin{array}{c}
                    I\otimes y^T\\
                    0           \\
                    0           
                \end{array}
                \Bigg|
                \begin{array}{c}
                    I \\
                    0 \\   
                    0
                \end{array}
                \Bigg|
                \begin{array}{c}
                    \lambda^T\otimes I \\
            		I\otimes y^T \\     
            		0                
                \end{array}
                \Bigg|
                \begin{array}{c}
                    0 \\
                    -I \\   
                    0
                \end{array}
                \Bigg|
                \begin{array}{c}
                    \nu^T\otimes I \\   
                    0 \\                
                    D(\nu)I\otimes y^T
                \end{array}
                \Bigg|
                \begin{array}{c}
                    0 \\
                    0 \\   
                    -I
                \end{array}
            \right],
	\end{aligned}
\end{align}
where $\otimes$ denotes the Kronecker product.  
%The resulting backpropagation expression is then obtained by direct application of~\eqref{eq:chain_rule} to \eqref{eq:JFyk1} and \eqref{eq:JFyk}. 

% \subsubsection{Avoid Explicit Jacobian}

Furthermore, we show that, as in the case of backpropagating explicit layers, explicit construction of $\left[ J_{F, y^{(k)}} \right]$ is avoidable.  Taking the partial derivatives of the loss with respect to $Q$ for example, observe in the following expression that the multiplication operation on $J_{F, Q}$ is replaced with a Kronecker product on vector $y^T$:
\begin{align}\label{eq:simplification}
	\begin{aligned}	
	\frac{\partial{L}}{\partial{Q}} &= \frac{\partial{L}}{\partial{y^{(k+1)}}} \left[ J_{F, y^{(k+1)}} \right]^{-1} \left[ J_{F, Q} \right] \\
	&= \frac{\partial{L}}{\partial{y}} \left[ J_{F, y^{(k+1)}} \right]^{-1} \left[ I\otimes y^T \right] \\
	&= \frac{\partial{L}}{\partial{y}} \left[ J_{F, y^{(k+1)}} \right]^{-1} \otimes y^T.
	\end{aligned}
\end{align}
% thus replaces thOe term $\left[ I\otimes y^T \right]$ of size $|Q|\times|y|$ with $y^T$ which is of size $|y|$.  
Other components of $\left[ J_{F, y^{(k)}} \right]$ can be simplified in a similar fashion.

\subsubsection{Automatic Differentiation of the QP Layer}

With the techniques introduced in~\ref{sec:auto_diff} we are even able to avoid to derive~\eqref{eq:JFyk1} and~\eqref{eq:JFyk} (as well as any attempt for simplification, e.g.,~\eqref{eq:simplification}) at all.  We introduce the following auxiliary variables: 
\begin{align}
    \begin{aligned}
        &z^{(k)} = \left[ y^{(k+1)} \Big| y^{(k)} \right] = [y, \lambda, \nu, Q, q, A, b, G, h]\\
        &z^{(k+1)} = \left[ z_1 \Big| z_2 \Big| z_3 \right] = 0,
    \end{aligned}
\end{align}
and thus obtain an explicit representation $F(z^{(k)})=z^{(k+1)}$ of the QP layer by rewriting~\eqref{eq:QP_KKT} as 
\begin{align}\label{eq:QP_KKT2}
F:
\left\{
	\begin{aligned}
		Qy + q + A^T\lambda +  G^T\nu &= z_1\\
		Ay - b &= z_2\\
		D(\nu)(Gy - h) &= z_3.
	\end{aligned}
	\right.
\end{align}
As $z_1$, $z_2$ and $z_3$ are effectively explicit functions of $z^{(k)}$, we are able to use existing automatic differentiation implementations to compute $\frac{\partial{z_1}}{\partial{z^{(k)}}}$, $\frac{\partial{z_2}}{\partial{z^{(k)}}}$ and $\frac{\partial{z_3}}{\partial{z^{(k)}}}$ without any manual derivation, which constitute 
\begin{align}
    \frac{\partial{z^{(k+1)}}}{\partial{z^{(k)}}} = \frac{\partial{F}}{\partial{z^{(k)}}} = \left[ J_{F, y^{(k+1)}} \Big| J_{F, y^{(k)}} \right].
\end{align}

The automatic differentiation feature of the implicit layer comes extremely handy when the expression of the functions is complex, as is shown in the example in Section~\ref{sec:showcase_graphmatch}.

\subsubsection{Experiments - QP Layers}
% \subsubsection{Experiment on MNIST}

\begin{figure}[ht]
	\centering
	\begin{subfigure}{0.48\textwidth}
		\centering
		\includegraphics[width=\linewidth]{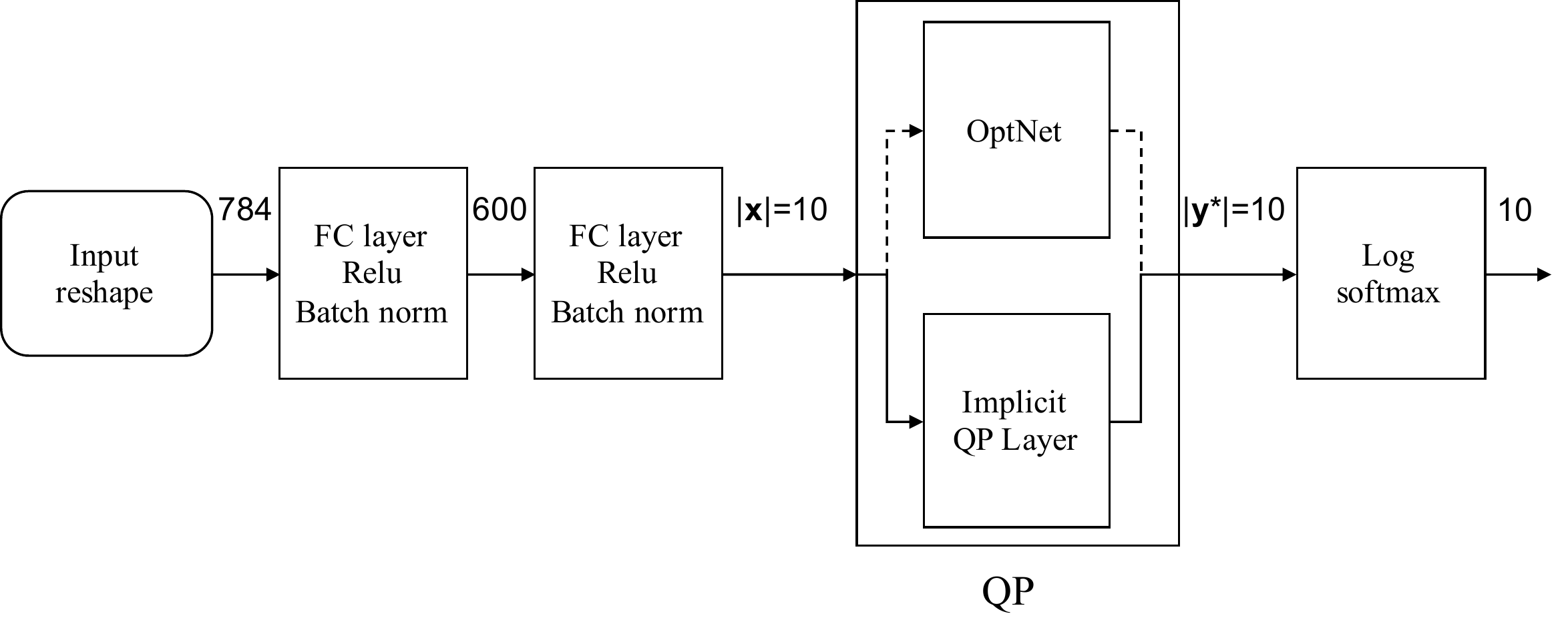}
	\end{subfigure}%
	\caption{The hand written digits recognition pipeline. The solid branch represents using an implicit layer to solve the QP problem and the dashed branch represents using OptNet.}
	\label{fig:mnist_pipeline}
\end{figure}

We verify the formulation of the QP layer on the task of hand digit recognition on MNIST.  The upper branch of figure~\ref{fig:mnist_pipeline} shows the pipeline.  Two fully connected layers take input from the vectorised $28\times28$ image, followed by a layer that solves a QP problem as defined by~\eqref{eq:QP}; the solution of the QP problem goes through softmax and yields the negative log likelihood loss.  We model this QP solving layer as an implicit layer.  

\begin{figure}[ht]
	\centering
	\begin{subfigure}[ht]{0.40\textwidth}
		\centering
		\includegraphics[width=\textwidth]{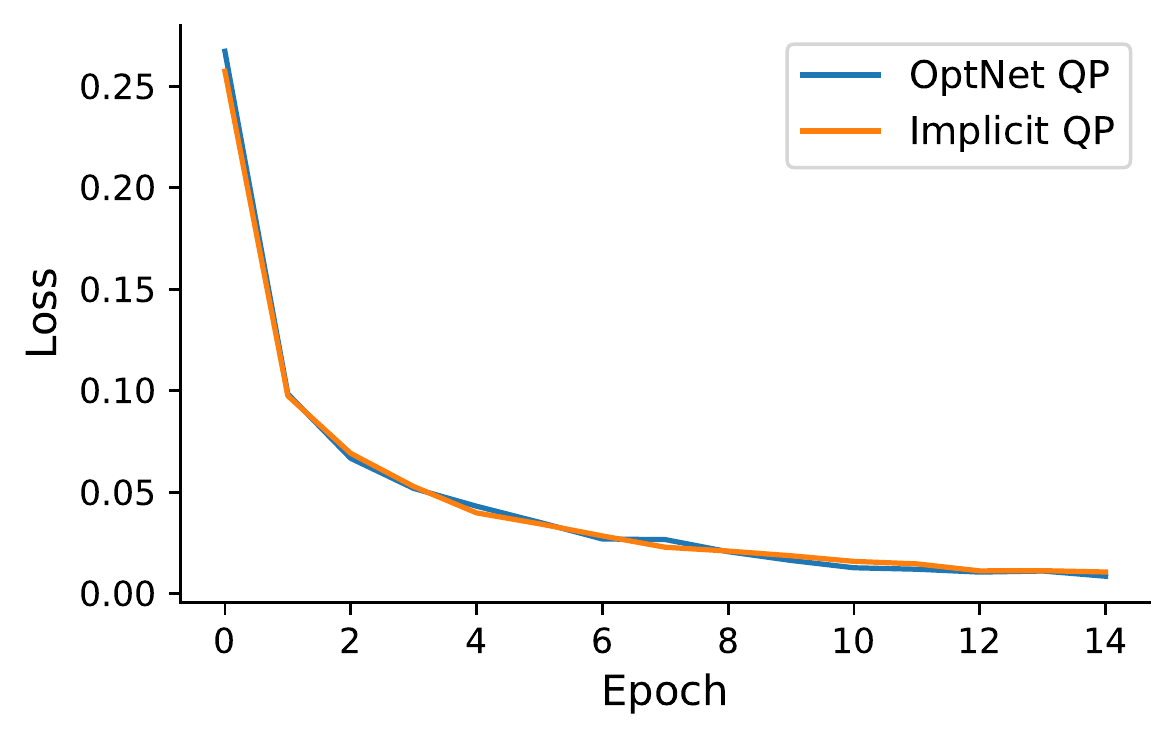}
	\end{subfigure}
	\vspace{1em}
	\begin{subfigure}[ht]{0.40\textwidth}
		\centering
		\includegraphics[width=\textwidth]{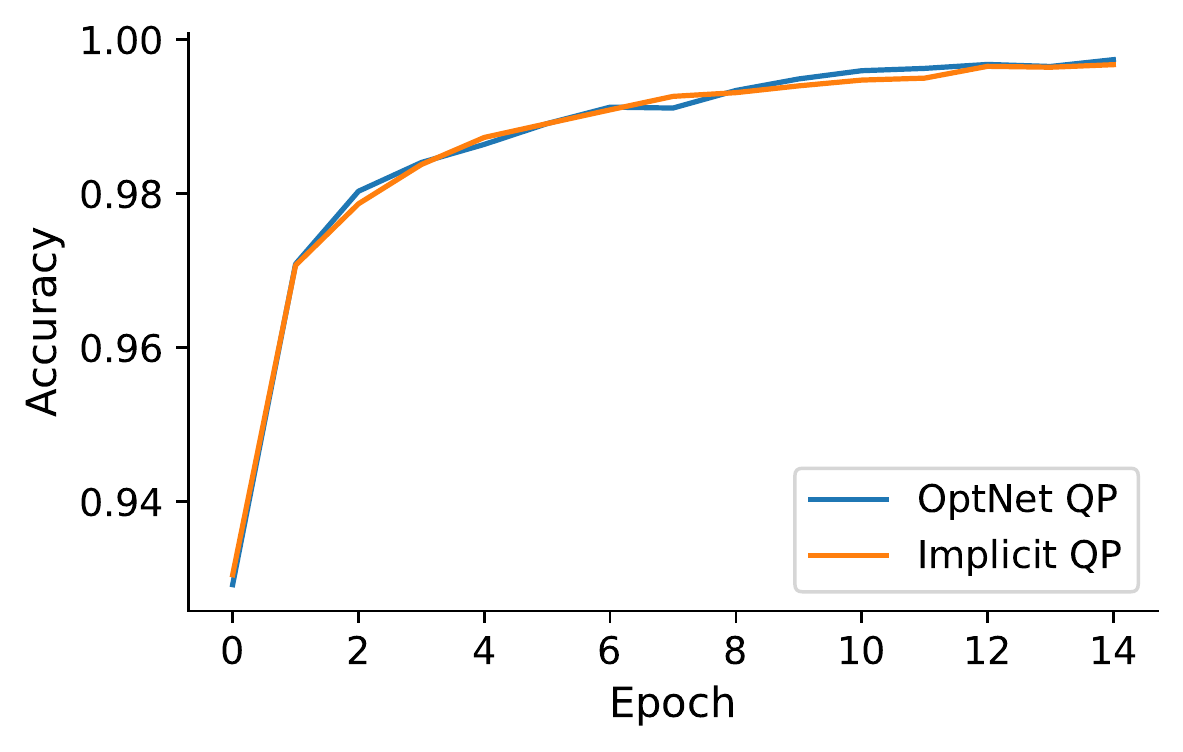}
	\end{subfigure}
	\caption{Hand written digits recognition learning curves.  Both pipelines achieves test accuracy above $98\%$ upon termination at 15 epochs.}
	\label{fig:mnist_loss_acc}
\end{figure}

The lower branch shows of figure~\ref{fig:mnist_pipeline} the pipeline designed by OptNet.  The only difference between the two pipelines is the component that solves the QP problem.  As for the choice of the solver of the forward pass, we adopt the same primal-dual interior point method.  This provides an equal comparison of the backward pass between the two pipelines.  
As expected, these two pipelines produce very similar convergence curves on 
both training and testing data, see figure~\ref{fig:mnist_loss_acc}, thus validating the 
efficacy of our proposed framework.
% As evidenced in figure~\ref{fig:mnist_loss_acc}, perhaps not surprisingly, the two pipelines produce essentially identical convergence curves.

The main purpose of this initial example application was to introduce the basic usage of 
implicitly defined layers and to demonstrate that its performance is comparable to 
existing state-of-the-art method . 
However, the key difference between these two methods is that 
\cite{amos2017optnet} is restricted to convex QPs and requires a manual implementation 
of the resulting backward pass, whereas our implicit framework can easily be adapted 
to a much broader class of layers. It also does not rely on dedicated backward pass 
implementations but can instead take advantage of existing automatic differentiation 
techniques to calculate the necessary partial derivatives. 

% Thus  far  we  have  introduced  basic  usages  of  implicit layer  and  shown  it  works well  on  modeling  QP  problem. However, the power of implicit layer does not end there.
% One key feature of the proposed framework lies in the level of generality.  As opposed 
% to OptNet which is bounded to solve QP problems, our implicit layer easily adapts to 
% other types of optimisation problems.  
% Next showcase on graph matching demonstrates how an implicit 
% layer models the Quadratic Programming with Quadratic Constraints (QCQP) problem and the 
% Maximizing a Rayleigh Quotient with Affine Constraints (MRQAC) 
% problem~\cite{cour2007balanced}. It also shows how the backward pass is independent on 
% the forward pass.

\subsection{Normalised Cuts Layers}

In this example we show how an implicit layer can be used to include a 
Normalised Cuts (NCut) framework~\cite{shi2000normalized} into an end-to-end 
trainable architecture. Normalised Cuts is a well established method for solving 
the perceptual grouping problem in computer vision. This is done by treating images as graphs and the image segmentation task as finding the cuts in such graphs that minimise the Normalized Cuts criterion. 

Given an image $I$ an undirected graph $G$, with vertices $V$ and edges $E$ where each vertex in $V$ corresponds to a single pixel in the image and the weights on the edges $E$ encodes the similarity between two pixels.  The non-negative weights of each such edge are represented by an affinity matrix $W$, with only non-negative entries and of full rank.  A Normalised Cut is then defined as the non-trivial partitioning of the graph $G$
into disjoint subsets $A$ and $B$ such that the criterion 
\begin{eqnarray}
\label{eq_ncut_def}
N_{cut}=\frac{cut(A,V)}{assoc(A,V)} +\frac{cut(B,V)}{assoc(B,V)}
\end{eqnarray}
is minimised.  Here $A\cup B=V$, $A\cap B=\emptyset$ and the normalizing term is defined as $assoc(A,V)=\sum_{i\in A, j\in V} w_{ij}$.
It is shown in  \cite{shi2000normalized} that % by relaxing \eqref{eq_ncut_def}
a continuous underestimator of the (minimal) Normalized Cut can be efficiently computed 
as the second smallest eigenvalue of the generalised eigensystem
\begin{align}
    \label{eq:ncut}
	L v =  \lambda D v,
\end{align}
where $L$ denotes the discrete Laplacian of the adjacency matrix $W$ of the image 
and $D$ is a diagonal matrix of the weighted graph order. 
%, see \cite{shi2000normalized} for further details. 

The implicit form of the NCut layer, defining the relationship between the 
resulting cut and the graph affinity, is obtained directly by rewriting~\eqref{eq:ncut} 
as 
% It involves little changes for the implicit layer that is to 
% model~\eqref{eq:ncut}.  
% With almost identical definition of underlying implicit function
\begin{align}
F:
\left\{
	\begin{aligned}
		(L - \lambda_2 D)v_2 = 0.\\
		v_2^Tv_2 -1=0,
	\end{aligned}
	\right.
\end{align}
Here $\lambda_2$ and $v_2$ denotes the second smallest generalised eigenvalue 
and corresponding eigenvector of~\eqref{eq:ncut}. 

\subsubsection{Experiments - NCut Layers}

\begin{figure}[ht]
	\centering
	\begin{subfigure}{0.48\textwidth}
		\includegraphics[width=\linewidth]{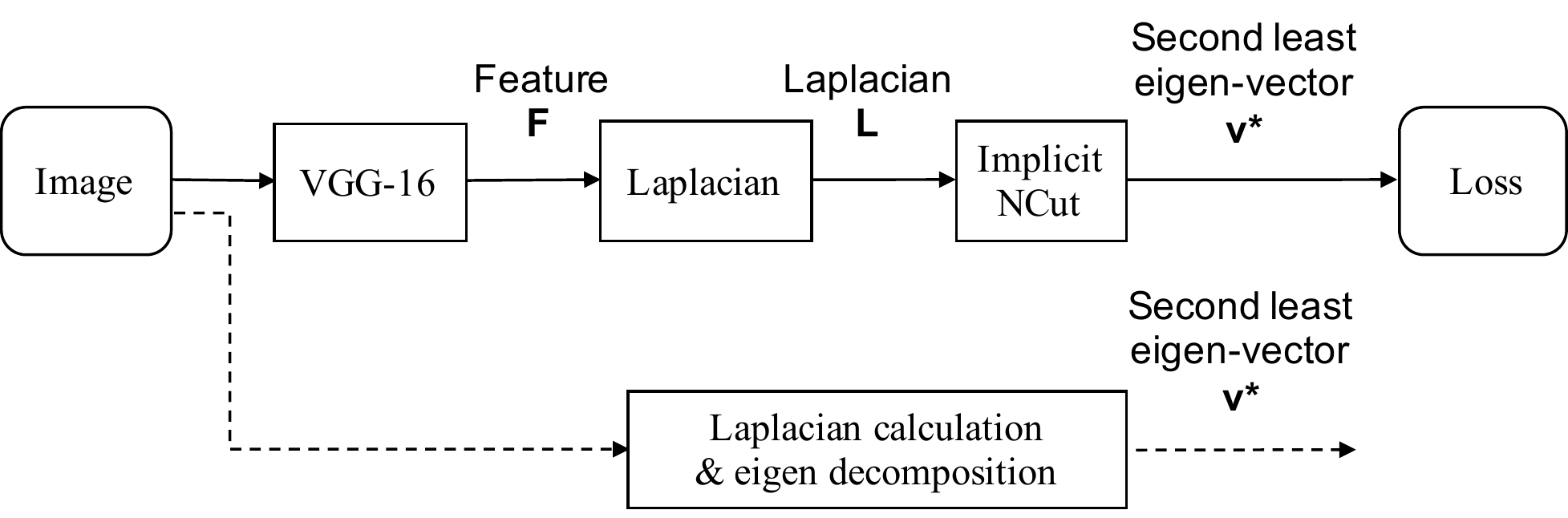}
	\end{subfigure}
	\caption{Image segmentation pipeline. The solid branch represents an end-to-end trainable pipeline.  It has an implicit NCut layer to decompose the Laplacian $L$, which is constructed based on learned features.  In contrast, the standard non-learning NCut method (the dashed branch) construct the Laplacian based on handcrafted features.}
	\label{fig:ncut_pipeline}
\end{figure}

The Normalised Cuts method is a graph theoretic formulation that aims to partition 
an image based on some measure of similarity between pixels (vertices) such that
similar pixels are grouped together and dissimilar ones may be separated.
This similarity measure is typically handcrafted, 
\cite{malik2001contourand,yu2003multiclass}. To evaluate the implicit Ncut layer 
we instead aim to learn this similarity metric from training data. 
The proposed implicit framework permits us to do so in an end-to-end fashion. 
We design a prototypical neural network with the NCut layer and compare it with the 
classical, non-learning based Normalised Cut method~\cite{shi2000normalized}, as shown 
in figure~\ref{fig:ncut_pipeline}. 

\begin{figure}[ht]
	\centering
	\begin{subfigure}{0.40\textwidth}
		\centering
		\includegraphics[width=\linewidth]{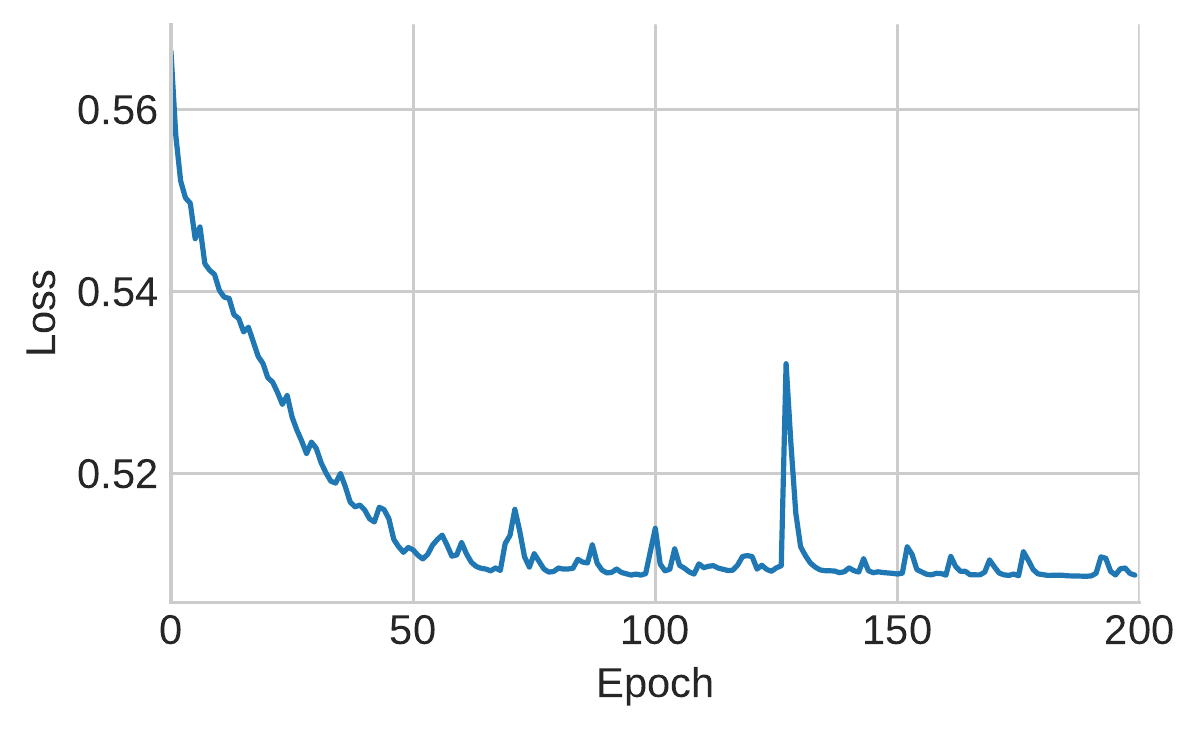}
		%\caption{}
	\end{subfigure}
	\vspace{1em}
	\begin{subfigure}{0.40\textwidth}
		\centering
		\includegraphics[width=\linewidth]{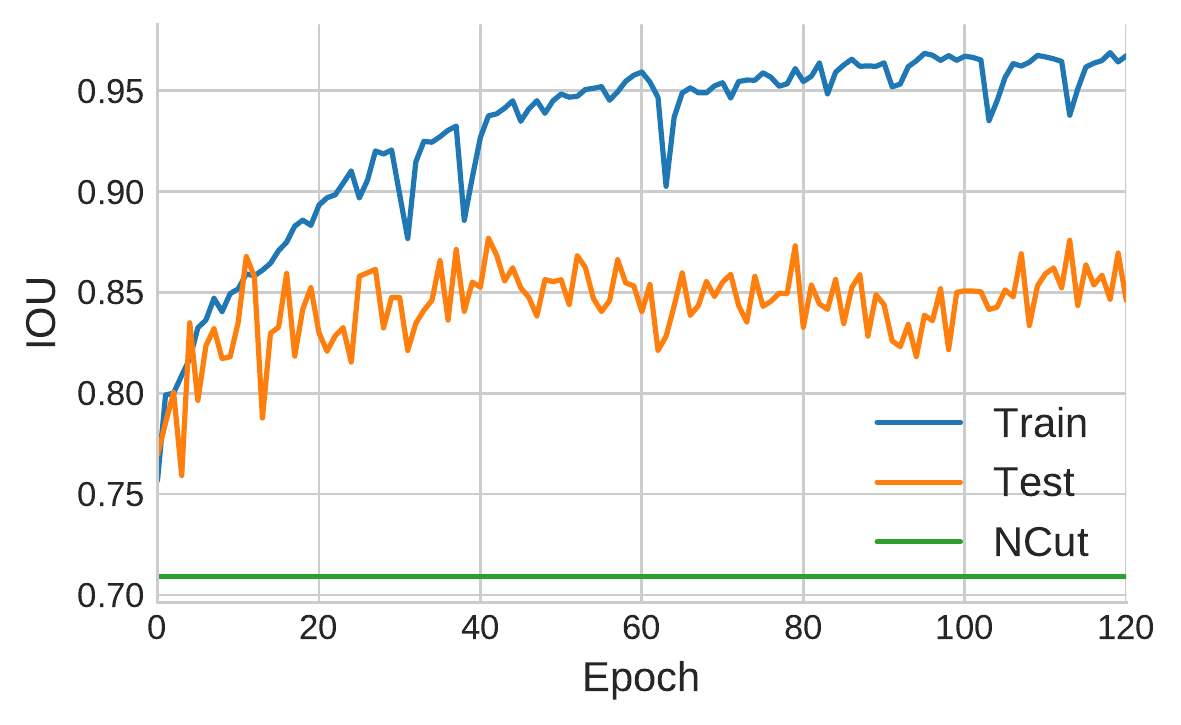}
		%\caption{}
	\end{subfigure}
	\caption{Convergence curve of the image segmentation pipeline.  The left panel plots the convergence of loss as training for the learning based branch. The right panel compares the IOU scores of the learning based and standard method.}
	\label{fig:hazysky_acc}
\end{figure}

We evaluate this pipeline %in figure~\ref{fig:ncut_pipeline} 
on the HazySky dataset~\cite{Song2018Sky} which contains 500 natural images with ground truth sky/non-sky segmentation mask.  Training data consists of 400 randomly sampled images with the rest for testing data.  Measured by the Intersection over Union (IOU) score, the accuracy of the learning based pipeline converges to just over$85\%$.  By comparison, the non-learning based NCut method only achieves an IOU score of about $70\%$. See figure~\ref{fig:hazysky_acc}. 

%\subsection{Level-set layer for surface reconstruction}
\subsection{Level Set Layers for Shape Inference}

Next we demonstrate how Level Sets, an established tool for 
numerical analysis of surfaces and shapes can be 
incorporated into current deep learning frameworks through an 
implicit formulation. We evaluate this approach on the task of inferring object 3D shapes from a 
single image. 
Representing shapes in end-to-end trainable neural networks has proven to be 
challenging task. A majority of existing learning-based approaches involving shapes or 
structures relying on either voxel occupancy 
\cite{girdhar2016learning,choy20163d,rezende2016unsupervised,richter2018matryoshka}, 
sparse point clouds \cite{qi2017pointnet,fan2017point} or explicit shape 
parameterisation \cite{liao2018deep}. 
Each of these representations comes with its own advantages and disadvantages, in 
particular for the application of shape inference in a learning framework.
Recent work \cite{Park_2019_CVPR,Michalkiewicz_2019_ICCV} has instead argued that \emph{Level Sets} constitute a more appropriate choice for the task of learned shape inference.

The Level Set method for representing moving interfaces was proposed independently 
by \cite{osher1988fronts} and \cite{dervieux1980finite}. 
This method defines a time dependent orientable surface $\Gamma(t)$ implicitly as the zero 
iso-contour, or level set, of a higher dimensional 
auxiliary scalar function, called the \emph{level set function}
or \emph{embedding function}, $\phi(x,t): \Omega \times \R \mapsto \R$,  as, 
\begin{align} 
\Gamma(t)= \left\{ x : \phi(x,t) = 0 \right\},
\label{eq:level_set}
\end{align}
with the convention that $\phi(x,t)$ is positive on the interior and 
negative on the exterior of $\Gamma$. 
The underlying idea of the level set method is to capture the motion of 
the iso-surface through the manipulation of the level set function $\phi$.

However, owing to this implicit definition of shape, existing deep learning frameworks 
can not incorporate this representation straightforwardly. Instead, the inference is 
either 
carried out on the embedding function $\phi$ \cite{Park_2019_CVPR} directly, rather than on the 
shape itself, thus resulting in suboptimal reconstructions, or by approximating metrics 
on the iso-surfaces of $\phi$ \cite{Michalkiewicz_2019_ICCV}. 

In this section we will show how the level set representations can be included exactly 
using implicit layers. Again, the aim here is not provide an exhaustive study of  
implicit representations of shape but rather to demonstrate the applicability of our 
proposed framework. 

The definition of the implicit layer that realises \eqref{eq:level_set} is provided 
directly from its definition as 
$0=\phi(x)$. 
However, here the input to this layer is a discrete representation of the embedding function $\phi$ 
and Theorem.~\ref{implicit_function_theorem} assumes a continuous function $F$. 
This can be accomplished by constructing a continuous surrogate of $\phi$ and using this to represent the desired implicit layer. We define 
\begin{align}
0= \phi_{tri}(y^{(k+1)};y^{(k)})
\end{align}
as our implicit layer representation of \eqref{eq:level_set}. 
Here $\phi_{tri}{\cdot;y^{(k)}}$ denotes the tri-linear interpolation of $y^{(k)}$ at $y^{(k+1)}$.

The forward pass through this layer can be obtained by any iso-surface extraction algorithm, 
see \cite{hansen2011visualization}. In this setup we use standard marching cubes \cite{lorensen1987marching}.

\subsubsection{Experiments - Level Set Layers}

\begin{figure}[h!]%[htb]
	\centering
% CARS
\includegraphics[width=.27\linewidth]{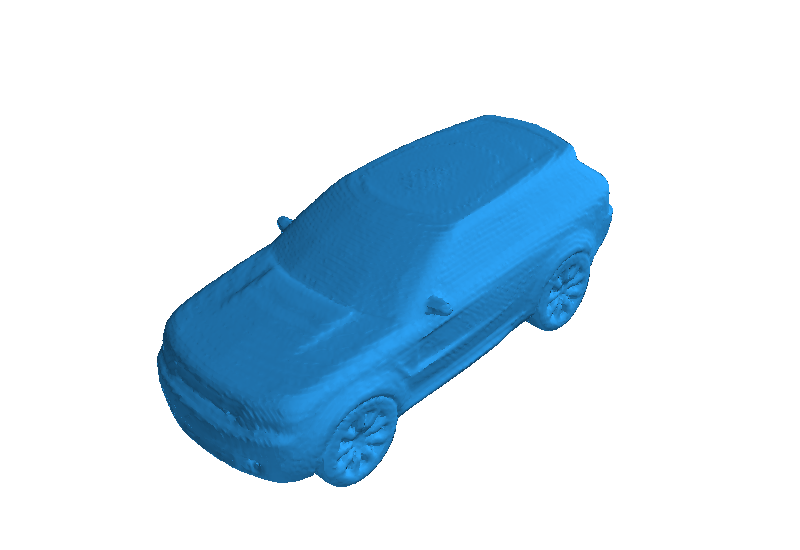}
\includegraphics[width=.16\linewidth]{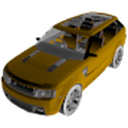}
\includegraphics[width=.27\linewidth]{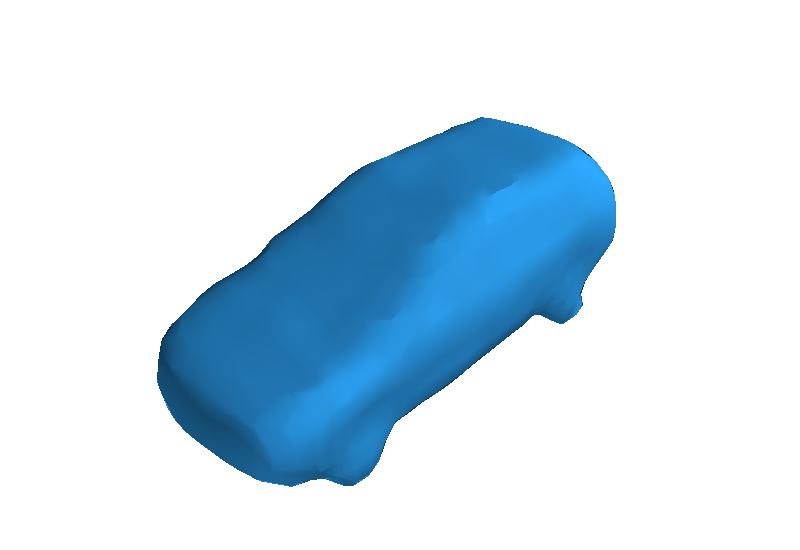}
\includegraphics[width=.27\linewidth]{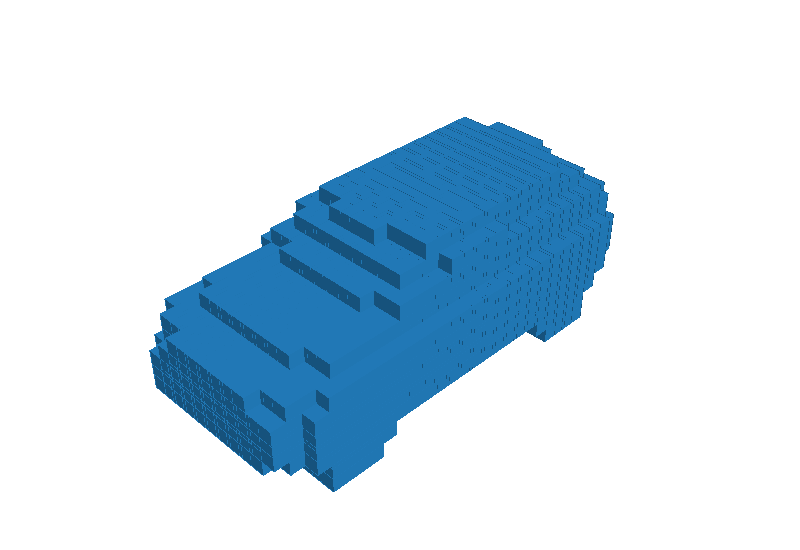}
\includegraphics[width=.27\linewidth]{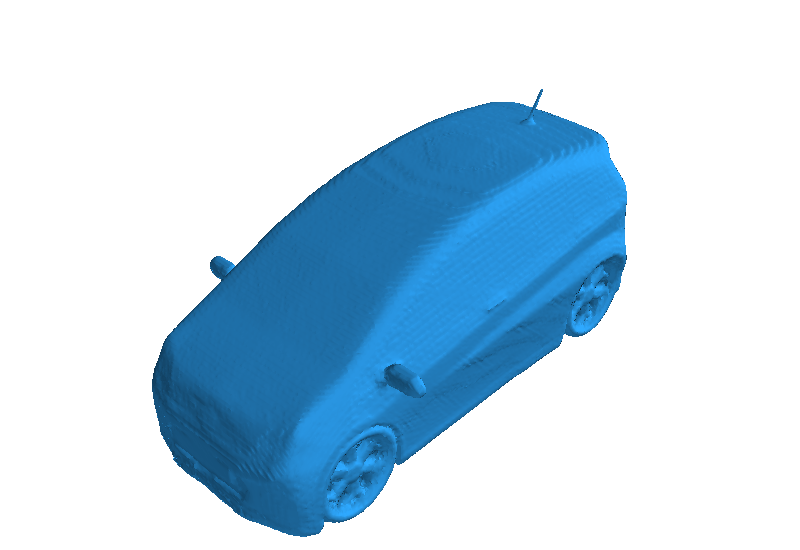}
\includegraphics[width=.16\linewidth]{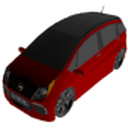}
\includegraphics[width=.27\linewidth]{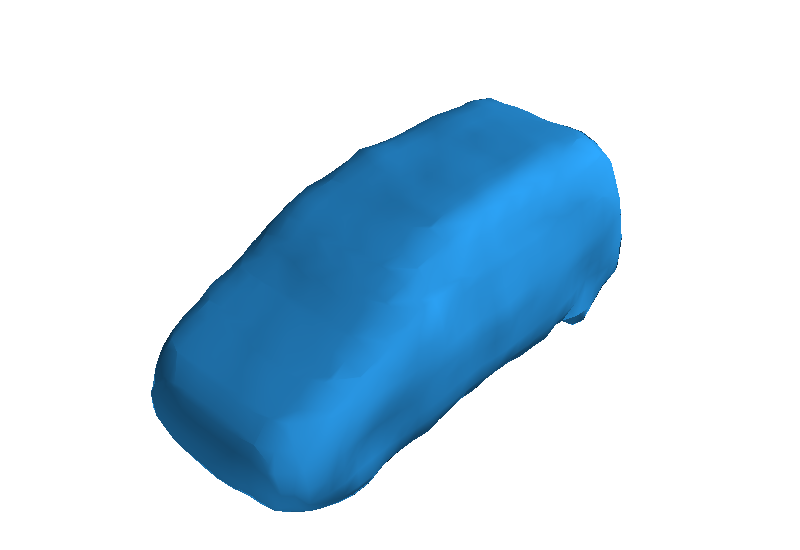}
\includegraphics[width=.27\linewidth]{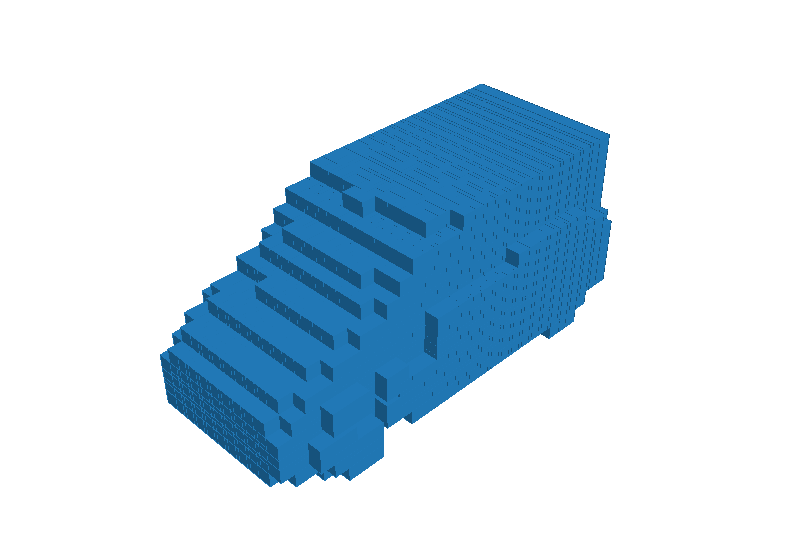}
\includegraphics[width=.27\linewidth]{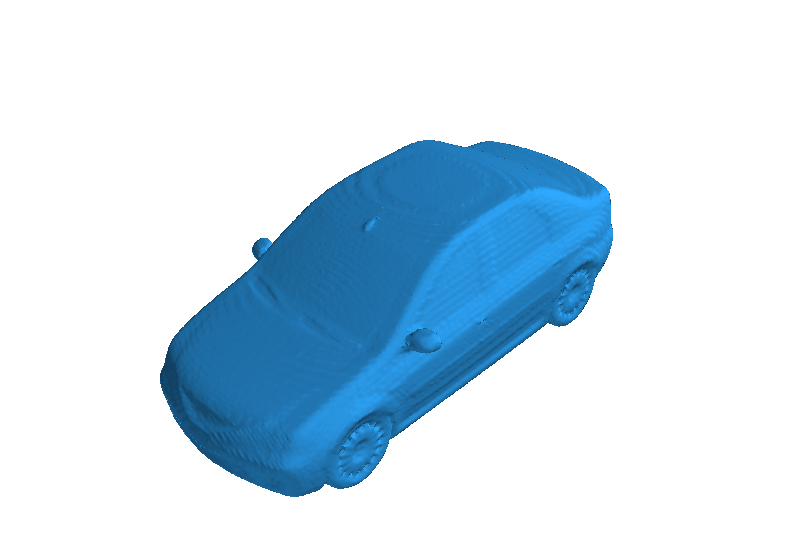}
\includegraphics[width=.16\linewidth]{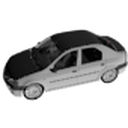}
\includegraphics[width=.27\linewidth]{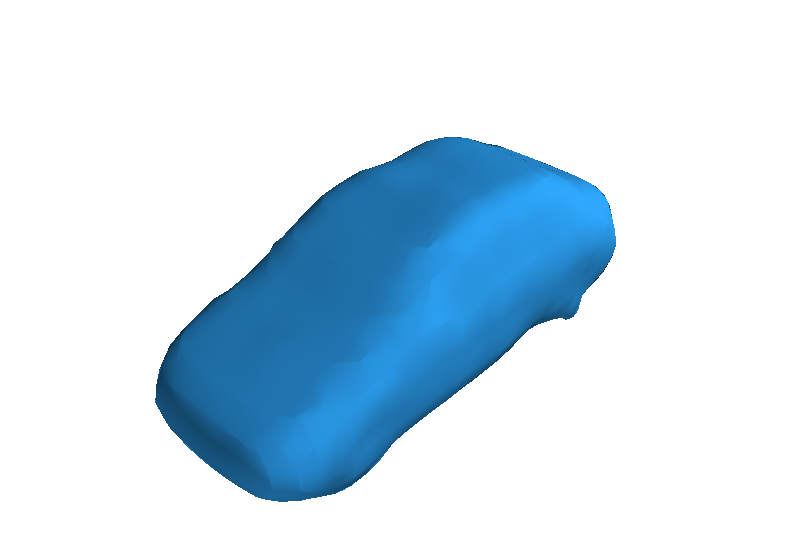}
\includegraphics[width=.27\linewidth]{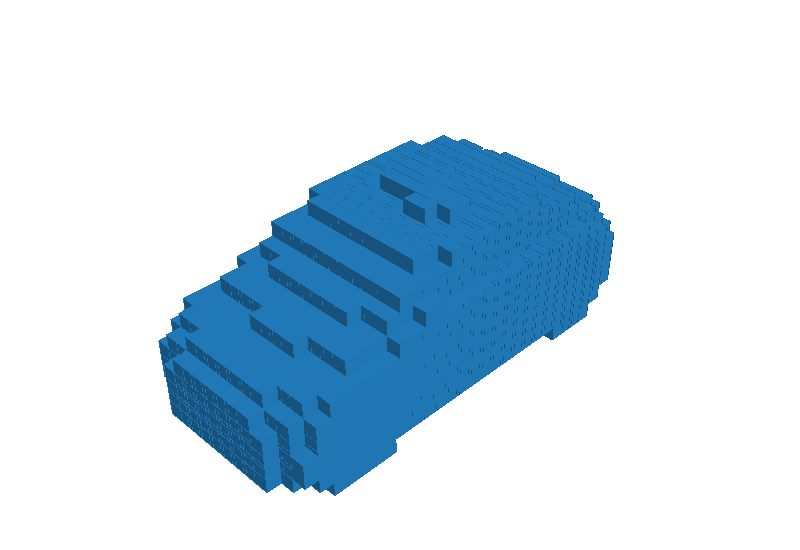}
%
% CHAIRS
\def\dd{.24}
\includegraphics[width=\dd\linewidth]{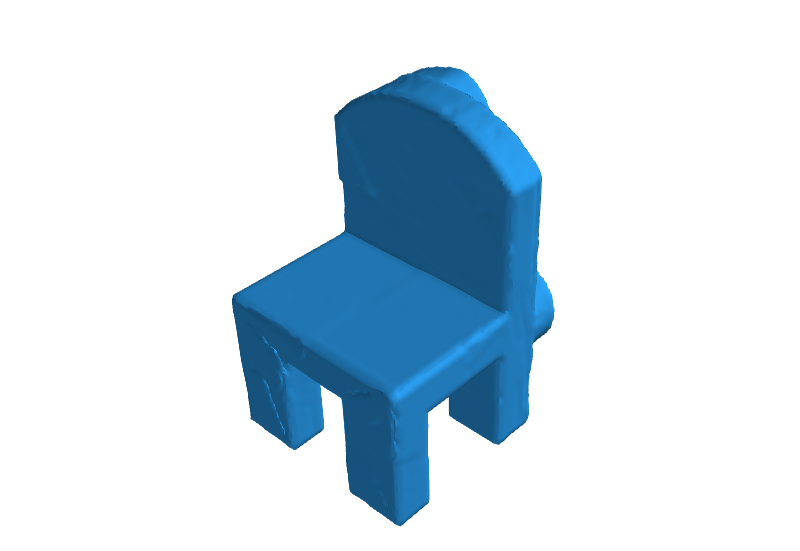}
\includegraphics[width=\dd\linewidth]{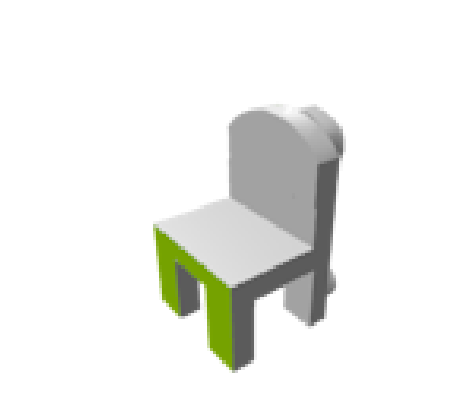}
\includegraphics[width=\dd\linewidth]{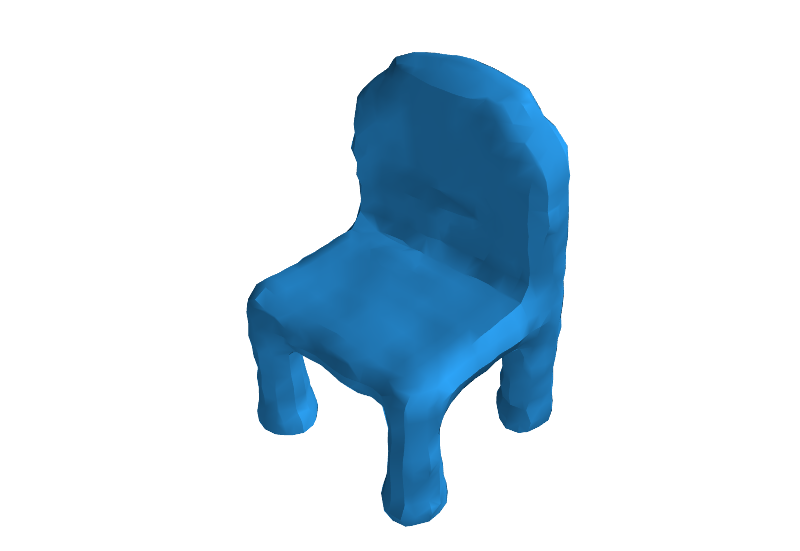}
\includegraphics[width=\dd\linewidth]{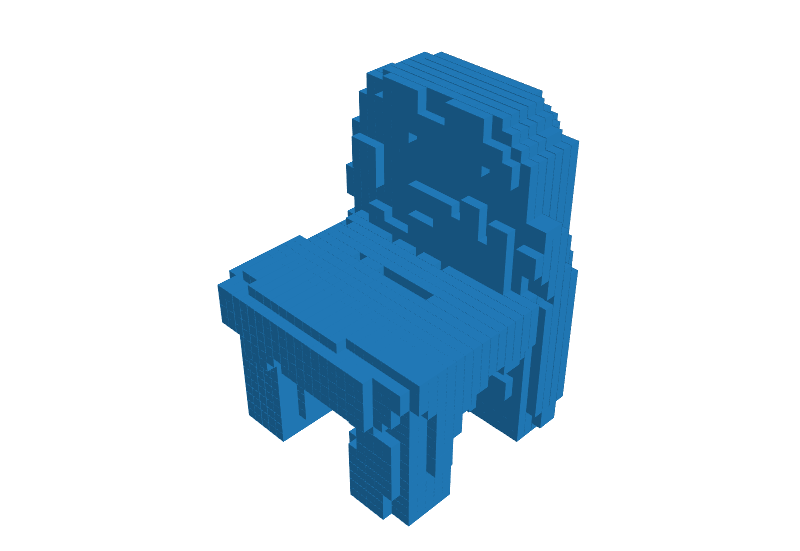}\\       
\includegraphics[width=\dd\linewidth]{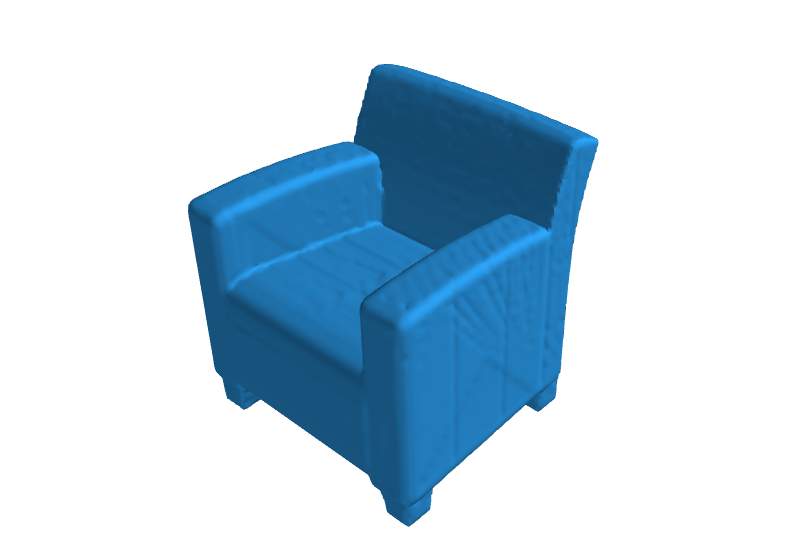} 
\includegraphics[width=\dd\linewidth]{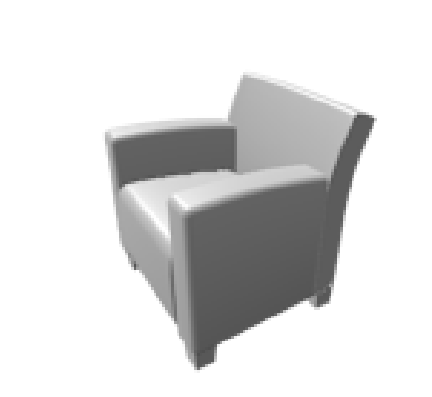} 
\includegraphics[width=\dd\linewidth]{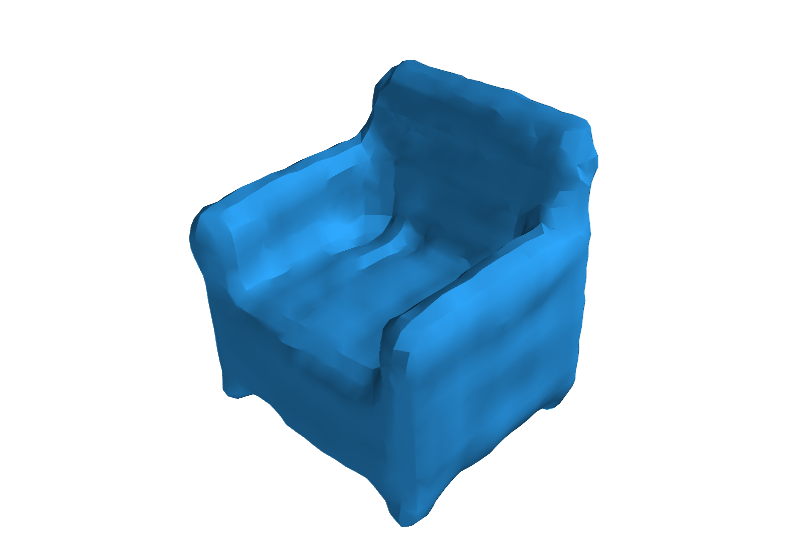} 
\includegraphics[width=\dd\linewidth]{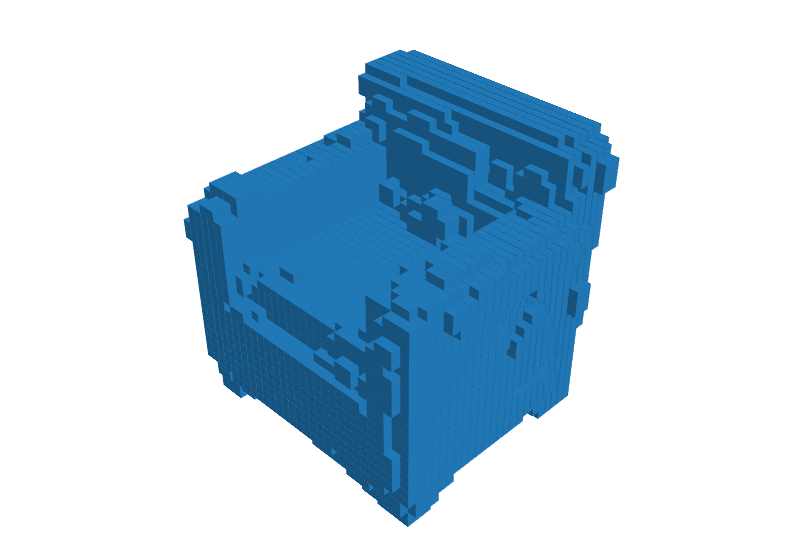}\\       
\includegraphics[width=\dd\linewidth]{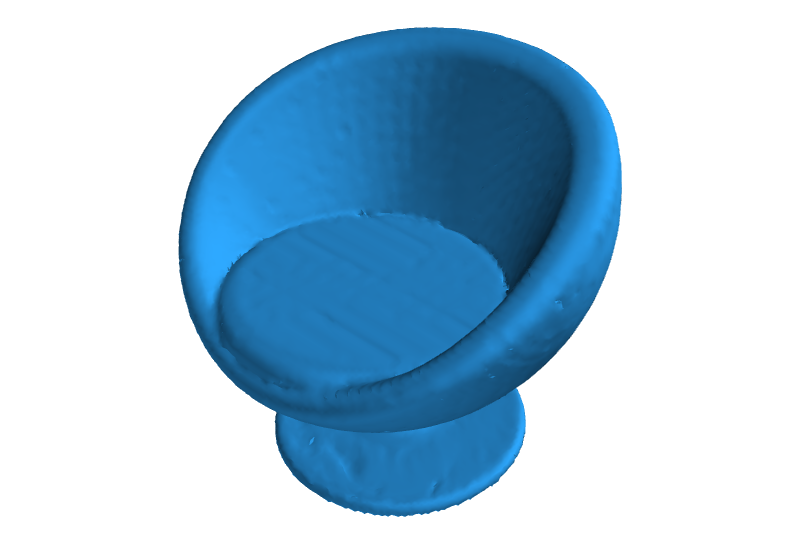} 
\includegraphics[width=\dd\linewidth]{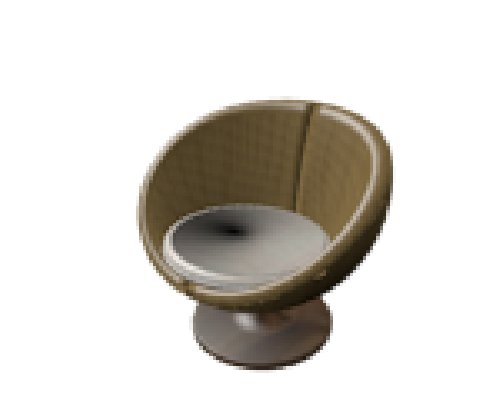} 
\includegraphics[width=\dd\linewidth]{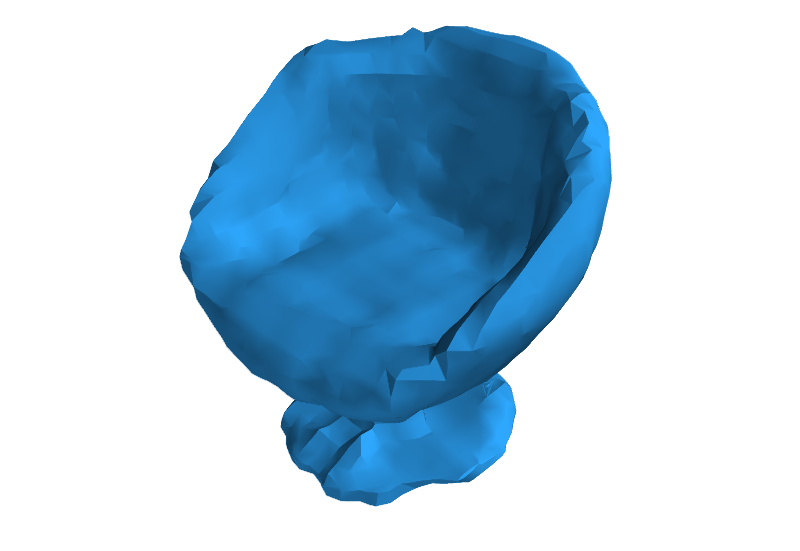} 
\includegraphics[width=\dd\linewidth]{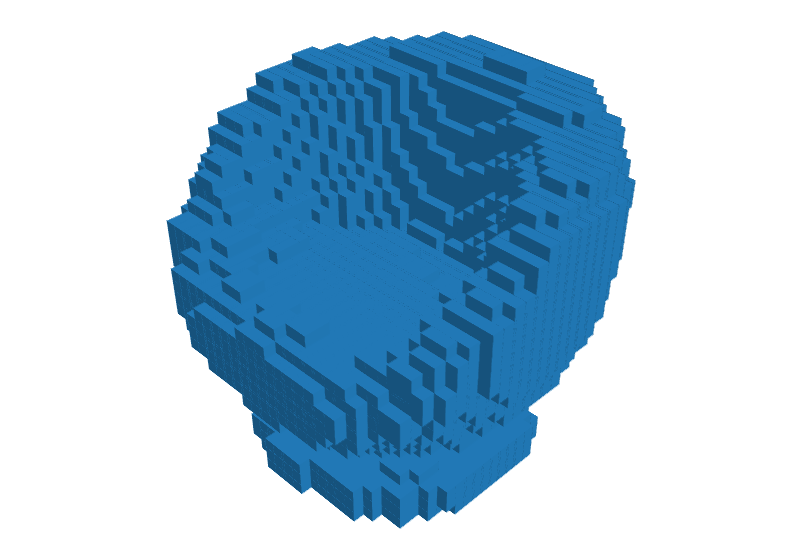}\\       
%
% Bottles
\def\dd{.24}
\includegraphics[width=\dd\linewidth]{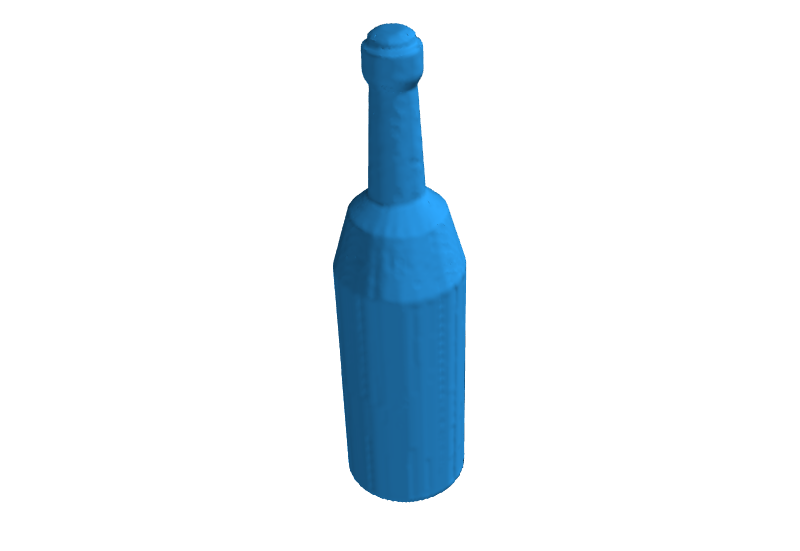} 
\includegraphics[width=\dd\linewidth]{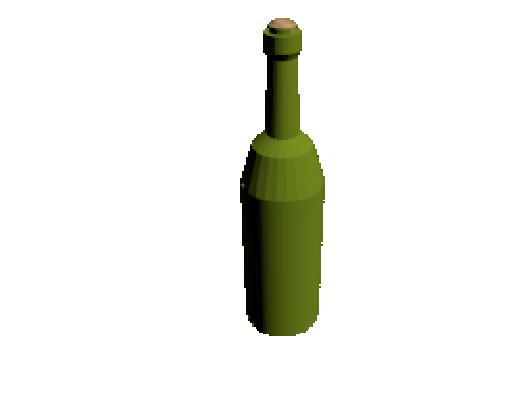}  
\includegraphics[width=\dd\linewidth]{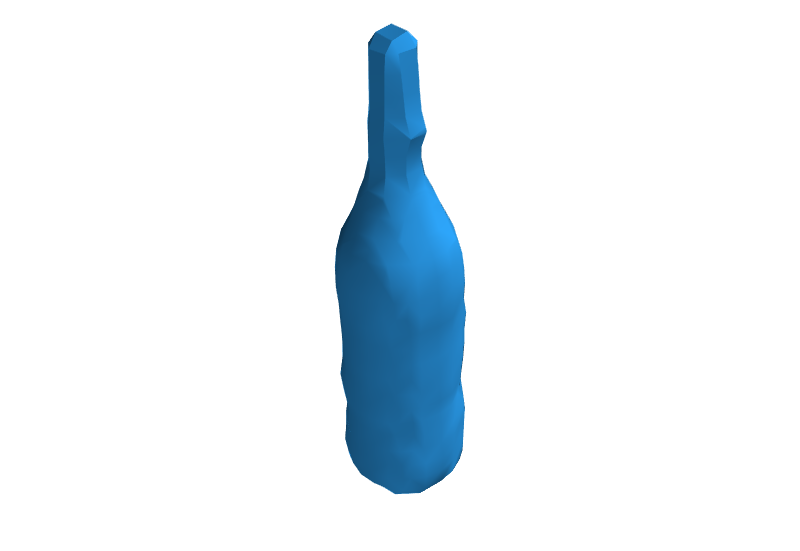} 
\includegraphics[width=\dd\linewidth]{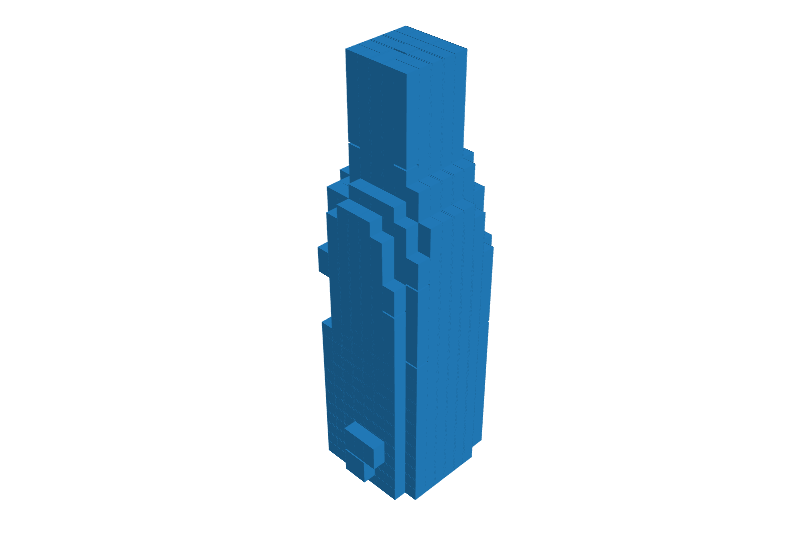}\\
\includegraphics[width=\dd\linewidth]{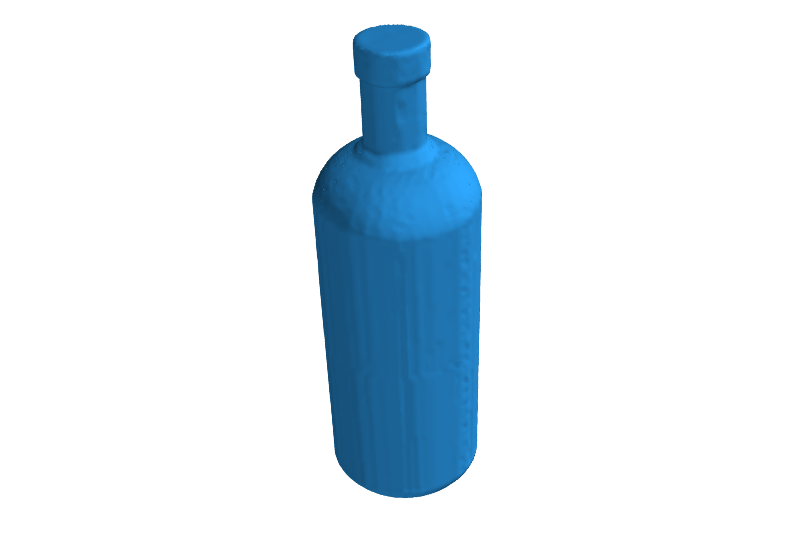}  
\includegraphics[width=\dd\linewidth]{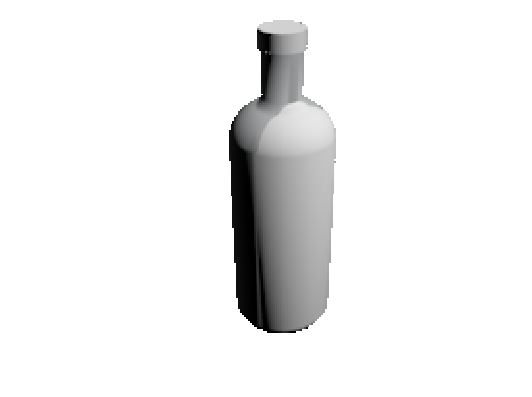}  
\includegraphics[width=\dd\linewidth]{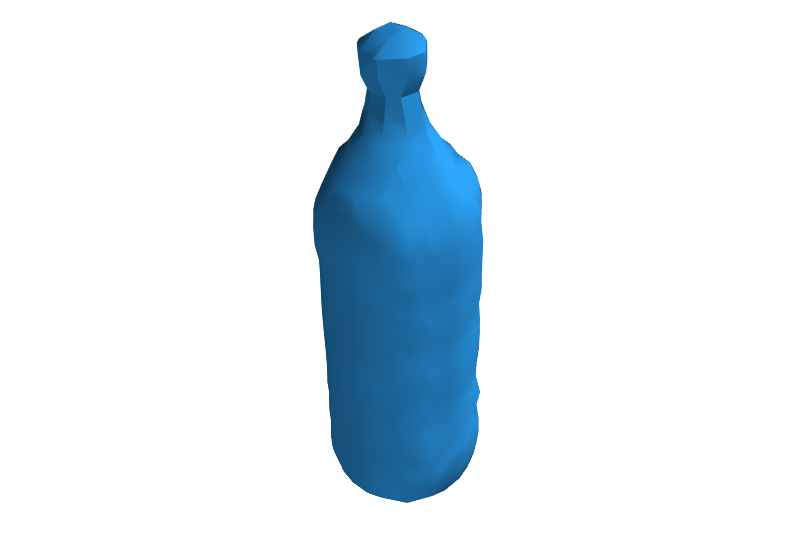} 
\includegraphics[width=\dd\linewidth]{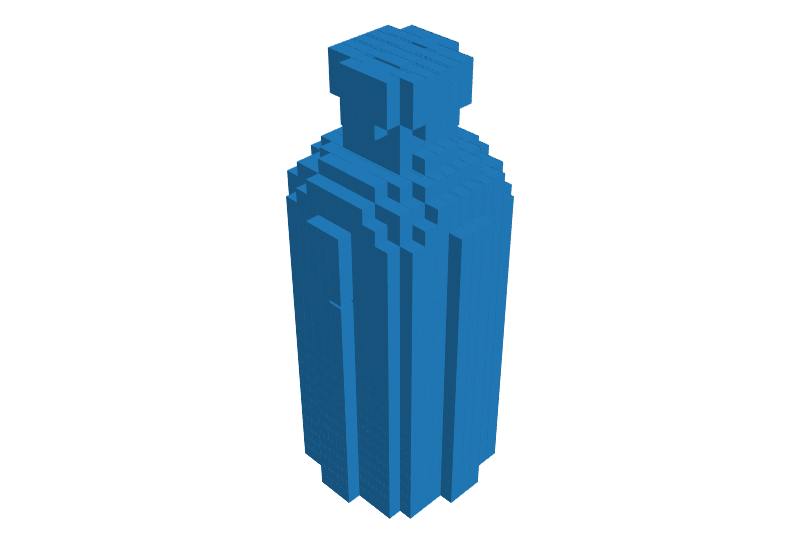}\\
\includegraphics[width=\dd\linewidth]{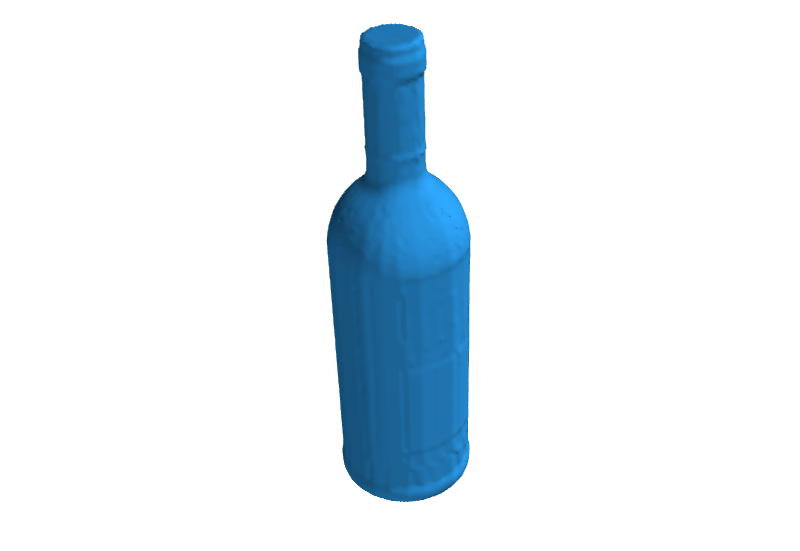}  
\includegraphics[width=\dd\linewidth]{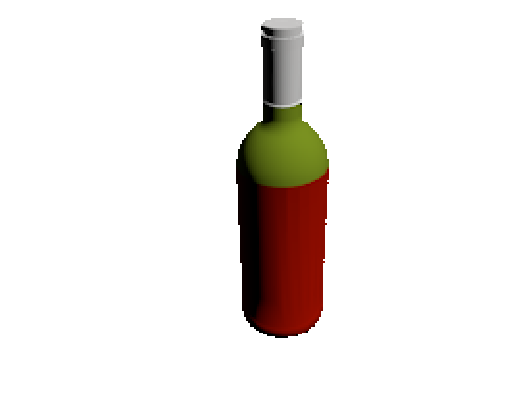}  
\includegraphics[width=\dd\linewidth]{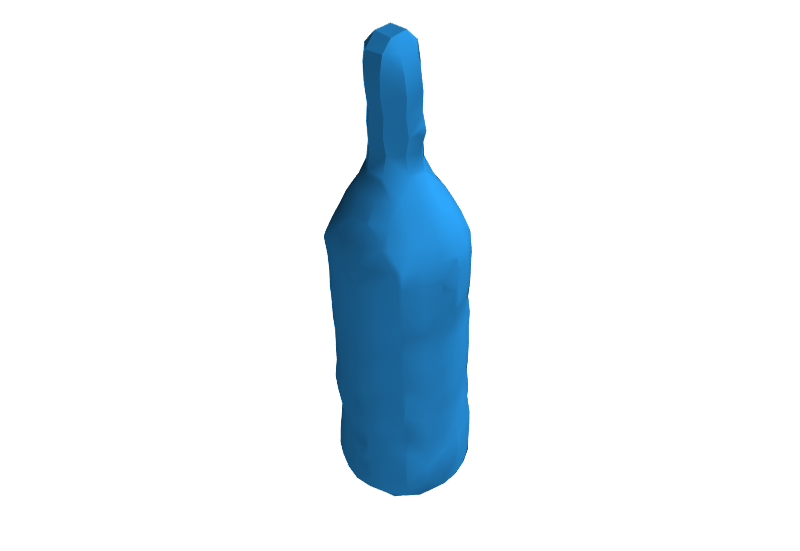} 
\includegraphics[width=\dd\linewidth]{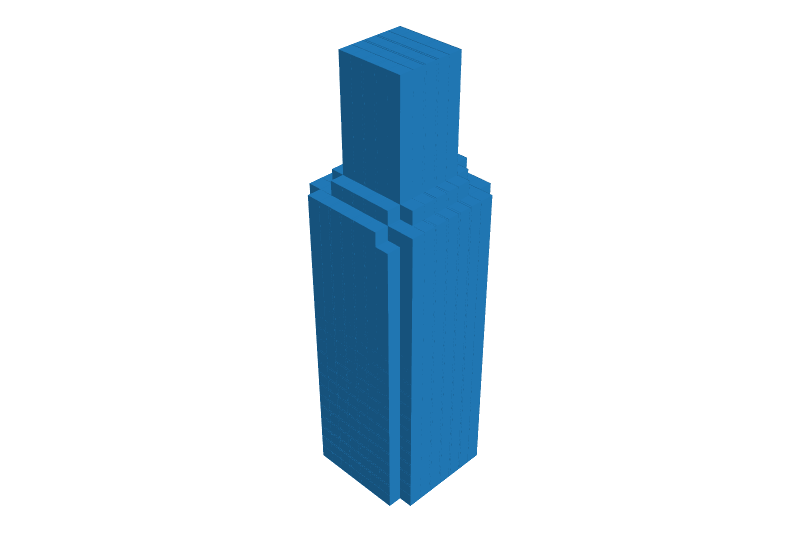}\\
\includegraphics[width=\dd\linewidth]{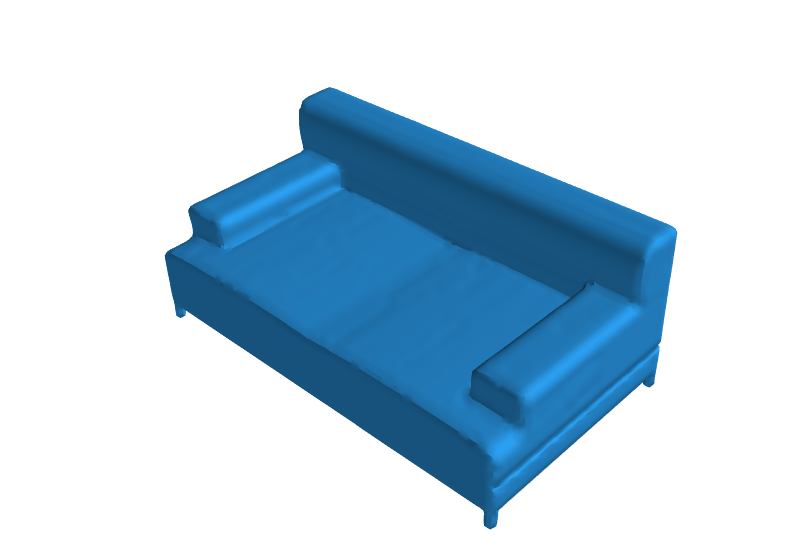}
\includegraphics[width=\dd\linewidth]{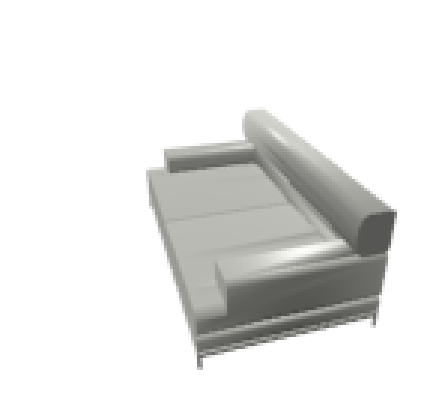} 
\includegraphics[width=\dd\linewidth]{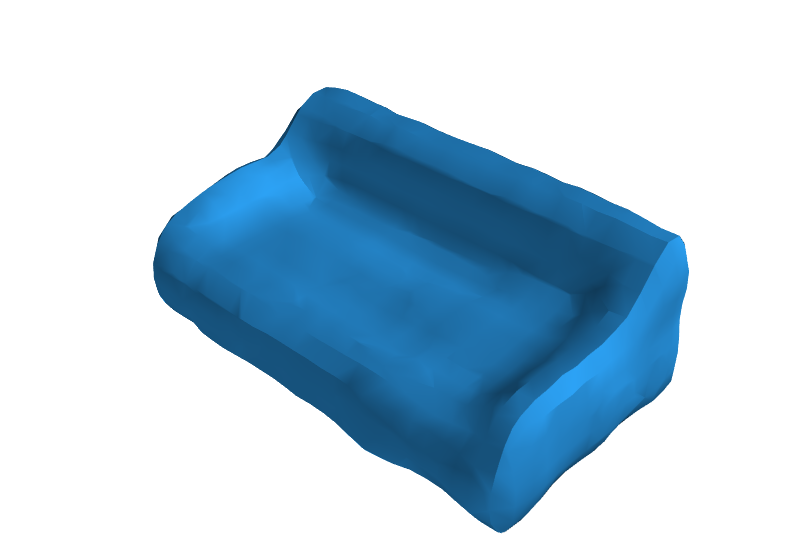} 
\includegraphics[width=\dd\linewidth]{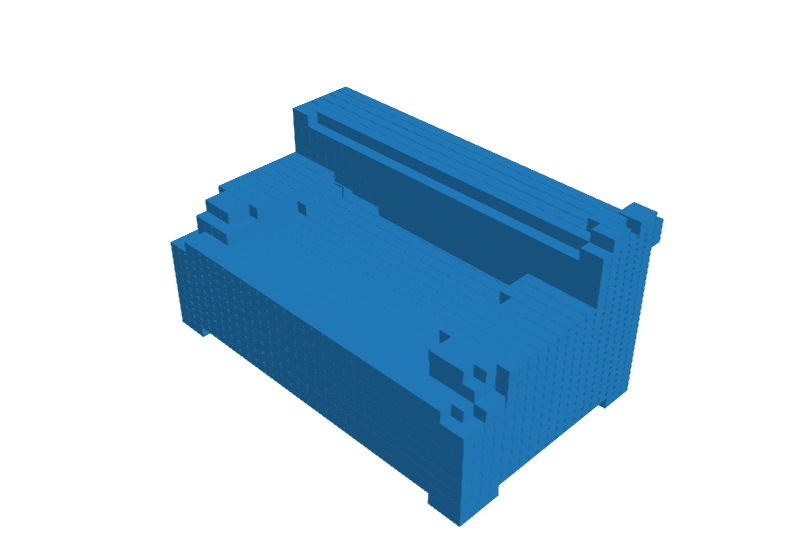}\\       
\includegraphics[width=\dd\linewidth]{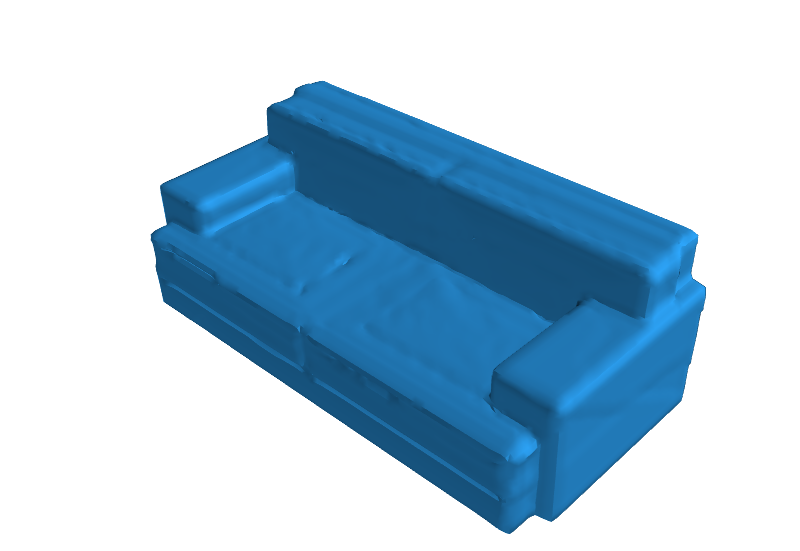} 
\includegraphics[width=\dd\linewidth]{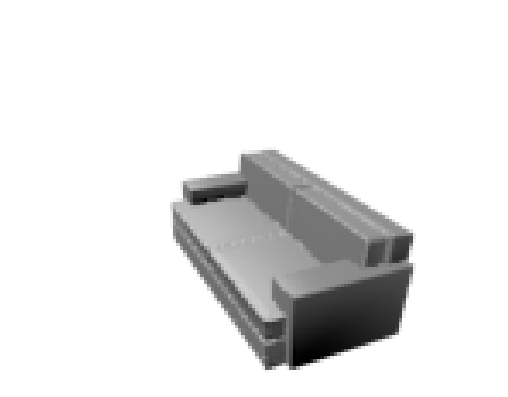}  
\includegraphics[width=\dd\linewidth]{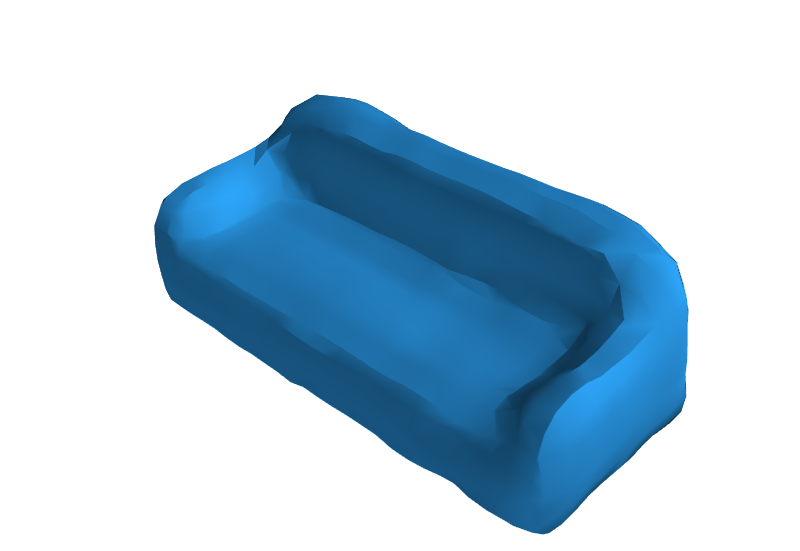} 
\includegraphics[width=\dd\linewidth]{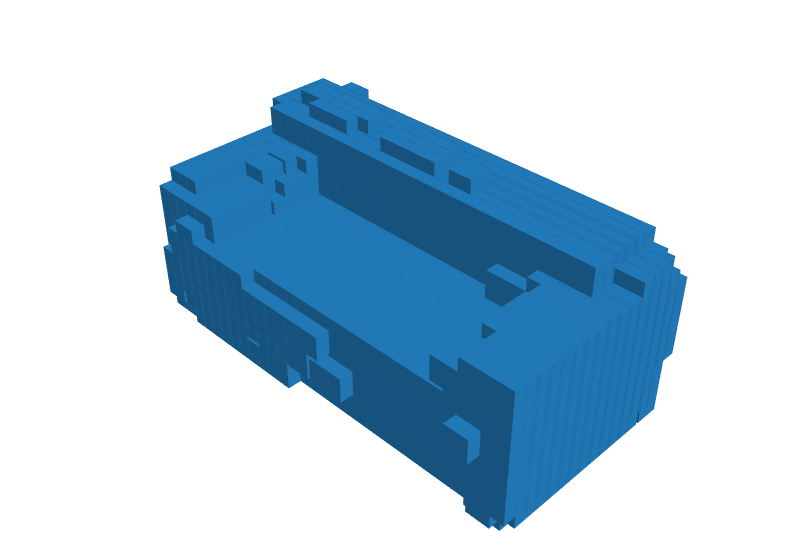}\\       
\includegraphics[width=\dd\linewidth]{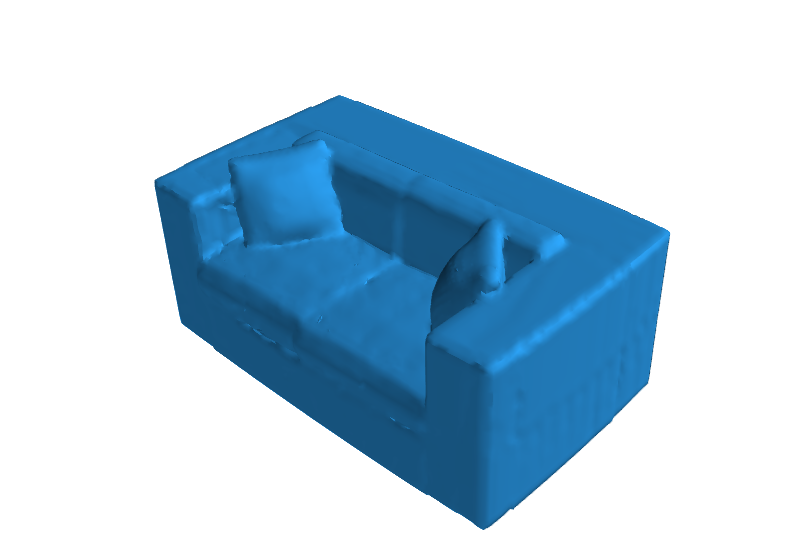} 
\includegraphics[width=\dd\linewidth]{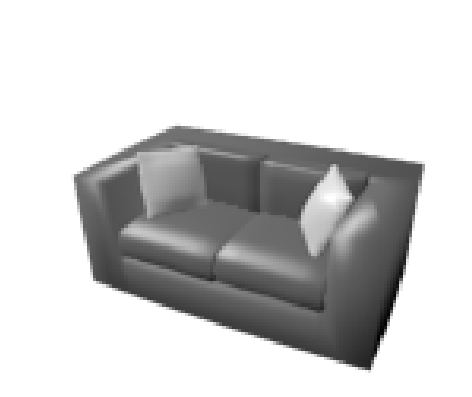} 
\includegraphics[width=\dd\linewidth]{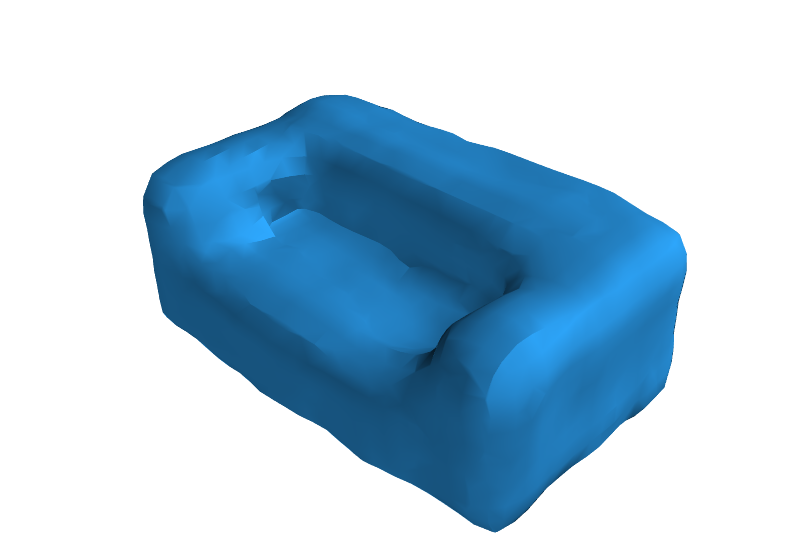} 
\includegraphics[width=\dd\linewidth]{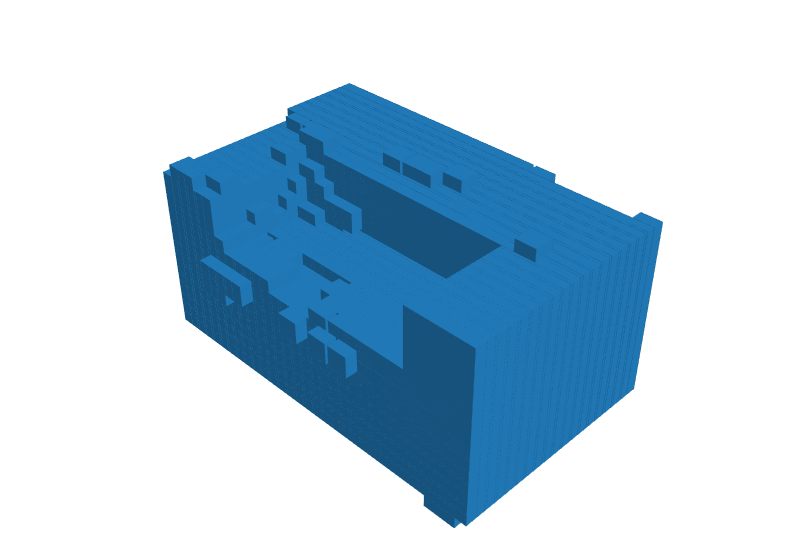}\\       
	\caption{Shape inference from a single image. Ground-truth (left), 
		input image (second column),
		level sets (third column),
		and voxels (right).
	}
	\label{fig:levelsets_cars}
\end{figure}

To evaluate this formulation we followed the implementation details provided in 
\cite{girdhar2016learning} and \cite{Michalkiewicz_2019_ICCV} as closely as possible with respect to preprocessing, image rendering and evaluation. 
Our network was evaluated with the proposed formulation on 8,000 3D models from four different  categories  
({'cars'}, {'chairs'}, {'bottles'} and {'sofas'}) in the ShapeNet dataset 
\cite{chang2015shapenet}. The results, using a $32^3$ 
resolution, are shown in figure \ref{fig:levelsets_cars} and table \ref{tab:table1}.
These results are comparable to, if not better than, those reported in 
\cite{Michalkiewicz_2019_ICCV}. 
Again the results are achieved with no dedicated implementation of the backward pass. 
\begin{table}
\begin{tabular}{@{}lrrrcrrrcrrrr@{}}
%\toprule
% && \multicolumn{2}{c}{{IoU}} & \phantom{a}&  \multicolumn{2}{c}{{Chamfer}} \\
&& \multicolumn{2}{c}{\textbf{IoU}} & \phantom{a}&  \multicolumn{2}{c}{\textbf{Chamfer}} \\
\cmidrule{3-4} \cmidrule{6-7} %\cmidrule{10-12}
\textbf{Category}& \phantom{ab} & Voxels & Level Sets &  &  Voxels & Level Sets  \\ 
% \cmidrule{1-1} \cmidrule{3-4} \cmidrule{6-7} %\cmidrule{10-12}
\cmidrule{1-7} 
%\\ \midrule
%
Car && 0.814  &  \textbf{0.868}  &   &   0.063 & \textbf{0.036}      \\ 
Chair && 0.100 &  \textbf{0.568}  &   &  \textbf{0.083}  & 0.089    \\ 
Bottle && 0.659 &  \textbf{0.782}  &   &  0.067 & \textbf{0.050}  \\ 
Sofa && 0.680 &  \textbf{0.737}  &  & 0.071    &  \textbf{0.056}  \\ 
% Telephone && 0.0963  &  \textbf{0.0510}  &   &  0.0658 & \textbf{0.0530}  \\ 
%\bottomrule
\end{tabular}
\caption{Average test errors using voxels and level sets representations.\label{tab:table1}}
\end{table}

% ----------------------------------------------------------------------------

\subsection{General argmin Layers - Graph Matching}
%\subsection{Graph Matching}
\label{sec:showcase_graphmatch}
As a final example we demonstrate how a general, non-linear, non-convex, constrained optimization 
problem can be attained through an implicit layer representation. The task we substantiate here is that of 
Graph Matching. This refers to the task of establishing correspondence between the nodes of two graphs based on similarity between nodes and edges.  
Incorporating graph matching into an end-to-end trainable neural network was first proposed in
~\cite{zanfir2018deep}. The solution proposed therein relied on approximate, fixed iteration algorithms
for solving the forward and backward pass through its graph matching layer. We here show what we believe
is a more efficient and elegant procedure for the same task. 

Graph matching is a fundamental combinatorial optimisation problems with a broad class of applications in vision.  It is usually formulated as a Quadratic Assignment Problem (QAP)~\cite{bazaraa1982use}, in which exists an affinity matrix $M$ that encodes unary (node-to-node) similarity and pairwise (edge-to-edge) similarity between the two graphs.

We use notations similar to what appeared in~\cite{zanfir2018deep} to formulate graph matching as a QAP.  Formally, given two graphs $\cG_1 = (V_1, E_1)$ and $\cG_2 = (V_2, E_2)$, with $|V_1|=n$ and $|V_2|=m$, let $y \in \{0,1\}^{nm}$ be an indicator vector such that $v_{ia} = 1$ if $i\in V_1$ is matched to $a \in V_2$ and $0$ otherwise.  We build an affinity matrix $M \in R^{nm\times nm}$ such that $M_{ia;jb}$ measures the similarity between edge $(i,j)\in E_1$ and edge $(a,b)\in E_2$, and the diagonal entries of $M$ measures node-to-node similarity.  The optimal assignment $y^*$ is obtained by solving the following QAP
\begin{align}\label{eq:QAP}
	\begin{aligned}
		&\underset{y}{\arg\!\max}~ y^TMy,\\
		&s.t.~ Cy = 1,~~C \in \{0,1\},~~y \in \{0,1\},
	\end{aligned}
\end{align}
where the binary matrix $C$ encodes one-to-one mapping constraints. General QAP problems are known to be NP-hard, so instead approximate solutions are typically considered for this class of problems. 

In this section we will study two different relaxed versions of \eqref{eq:QAP}. 
Firstly, the non-convex Quadratic Constrained Quadratic Programming (QCQP) problem obtained by dropping the constraint
$Cy = 1$ altogether and relaxing the binary constraint $y \in \{0,1\}$ to $||y||_2=1$:
\begin{align}\label{eq:rQAP_SM}
	\begin{aligned}
		&\underset{y}{\arg\!\max}~ y^TMy,\\
		&s.t.~ \|y\|_2 = 1.
	\end{aligned}
\end{align}
Even though \eqref{eq:rQAP_SM} is a nonconvex problem there  exist efficient solvers for finding 
the global minima of \eqref{eq:rQAP_SM}, and for this relaxation we use the Spectral Matching (SM) 
algorithm~\cite{cour2007balanced}. 

The implicit form of a graph matching layer is given by the KKT-conditions of \eqref{eq:rQAP_SM}
\begin{align}\label{eq:SM_KKT}
F:
\left\{
	\begin{aligned}
		& My + \lambda Iy=0\\
		& y^{T}y-1=0.
	\end{aligned}
	\right.
\end{align}
Our proposed implicit layer solves \eqref{eq:rQAP_SM} directly using SM algorithm followed by a 
bistochastic rounding to enforce the constraints $Cy = 1$ and $y \in \{0,1\}$. 
This constitutes the forward pass of the QCQP relaxed graph matching layer.
The backward pass is then obtained by applying section~\ref{sec:prop_through_imp_layers} to 
the above implicit form. 

The method proposed in ~\cite{zanfir2018deep} was also based on the above relaxation. However, 
the authors did not solve \eqref{eq:rQAP_SM} using standard SM but instead by carrying out a 
fixed number of Power Iterations (PI). As this approach is not guaranteed to solve \eqref{eq:rQAP_SM}
globally and owing to the potentially known poor convergence rate of PI,
the overall performance of this algorithm is unclear. 

A tighter relaxation of~\eqref{eq:QAP} is given by only relaxing the binary constraints on $y$ to 
$||y||_2=1$, yielding the following non-convex, constrained problem
\begin{align}\label{eq:QAP_SMAC}
	\begin{aligned}
		&\underset{y}{\arg\!\max}~ \frac{y^TMy}{y^Ty},\\
		&s.t.~ Cy = 1, ~y \geq 0.
	\end{aligned}
\end{align}
As with the previous relaxation, despite not being a convex problem, the above problem can also 
be solved efficiently with global optimality. Here we use the Spectral Matching with Affine Constraints 
(SMAC) algorithm from~\cite{cour2007balanced}. The backward pass is obtained through the implicit
formulation of ~\eqref{eq:QAP_SMAC}, which is given by
\begin{align}\label{eq:SMAC_KKT}
F:
\left\{
	\begin{aligned}
		& 2\frac{My^*y^{*T}y^* - y^{*T}My^*Iy^*}{(y^{*T}y^*)^2} + C^T\lambda^* - I\nu^*=0,\\
		& Cy^*-1=0,\\
		& D(\nu^*)Iy^*=0
	\end{aligned}
	\right.
\end{align}
This aproach guarantees doubly-stochastic output and exhibits better robustness against noise and 
outliers~\cite{zhou2012factorized}.
Note that the above relaxation cannot be solved using the approach of ~\cite{zanfir2018deep}. 
We in addition point out that the automatic differentiation feature of implicit layers, as described in Section~\ref{sec:auto_diff}, obviates the necessity of deriving symbolic expressions of derivatives for such a complex system of functions as~\eqref{eq:SMAC_KKT}, thus greatly simplifies actually implementations.

\subsubsection{Evaluation on Graph Matching}

\begin{figure}[htbp]
	\centering
	\begin{subfigure}{0.45\textwidth}
		\centering
		\includegraphics[width=\linewidth]{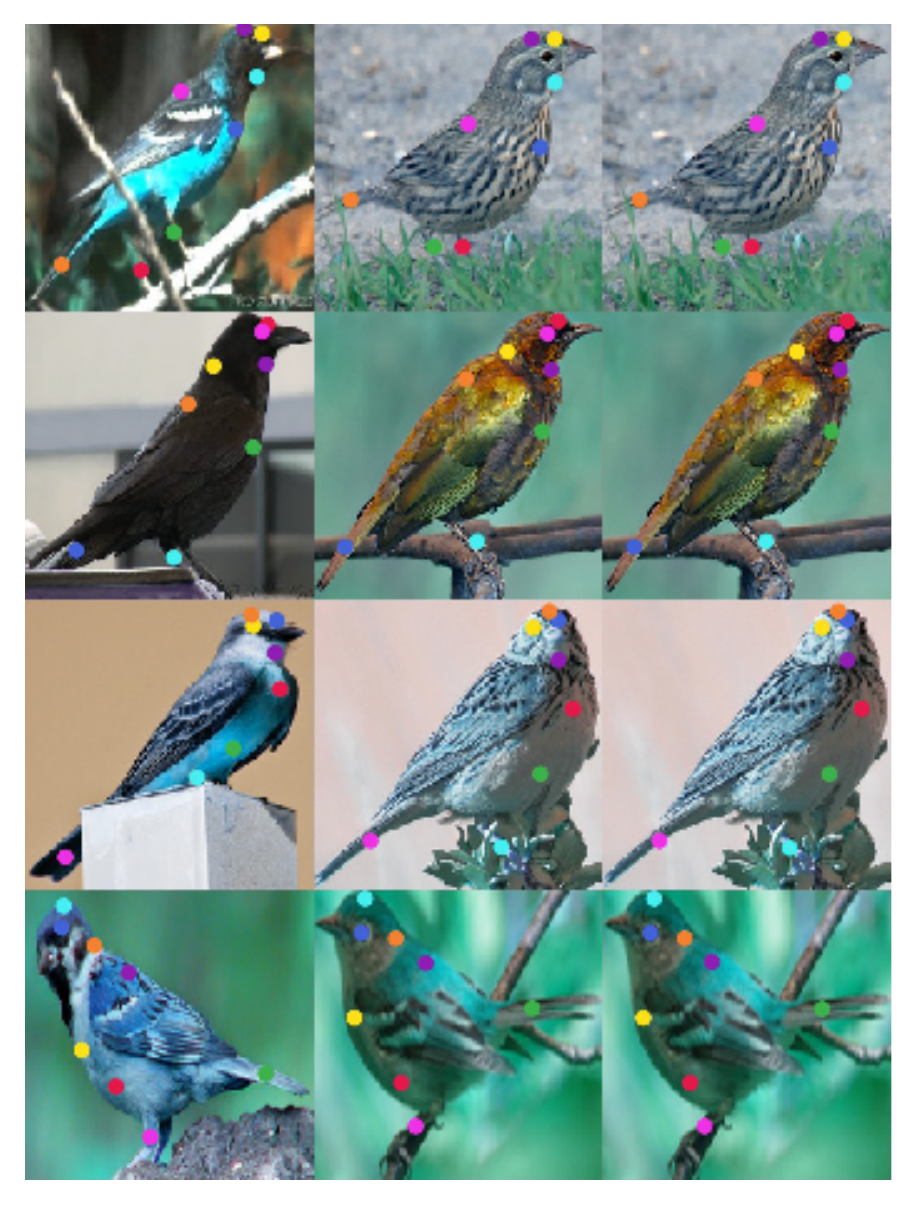}
	\end{subfigure}
	\caption{Qualitative results of the pipeline with an SMAC layer.  Landmarks are color coded.  The source image is on the left, the target image with predicted landmarks is in the middle, and on the right is the target image with ground truth landmarks.}
	\label{fig:qap_qualitative}
\end{figure}

We end this showcase with an experiment on the CUB-200-2011 dataset~\cite{WahCUB_200_2011}.  The dataset contains 11,788 images of 200 bird species, with a total of 15 semantic landmarks annotated by pixel location and visibility indicator.  A neural network is introduced to perform graph matching in order to establish an assignment that matches landmarks of the source image to those of the target image, see figure~\ref{fig:qap_qualitative}.  Since the purpose of this experiment is to verify the prototype of implicit layer rather than to show a comprehensive competition with other works, we simplify the task to operate on a subset of the CUB-200-2011 dataset that was built by~\cite{kanazawa2016warpnet}, which contains 5,000 images pairs with more than 50,000 ground truth matches.  Training dataset and test dataset are set at a ratio of 9:1.  We match fixed 8 randomly selected landmarks across images instead of matching up to 15 landmarks to avoid dealing with invisible landmarks.  

\begin{figure*}[ht]
	\centering
	\begin{subfigure}{0.75\textwidth}
		\centering
		\includegraphics[width=\linewidth]{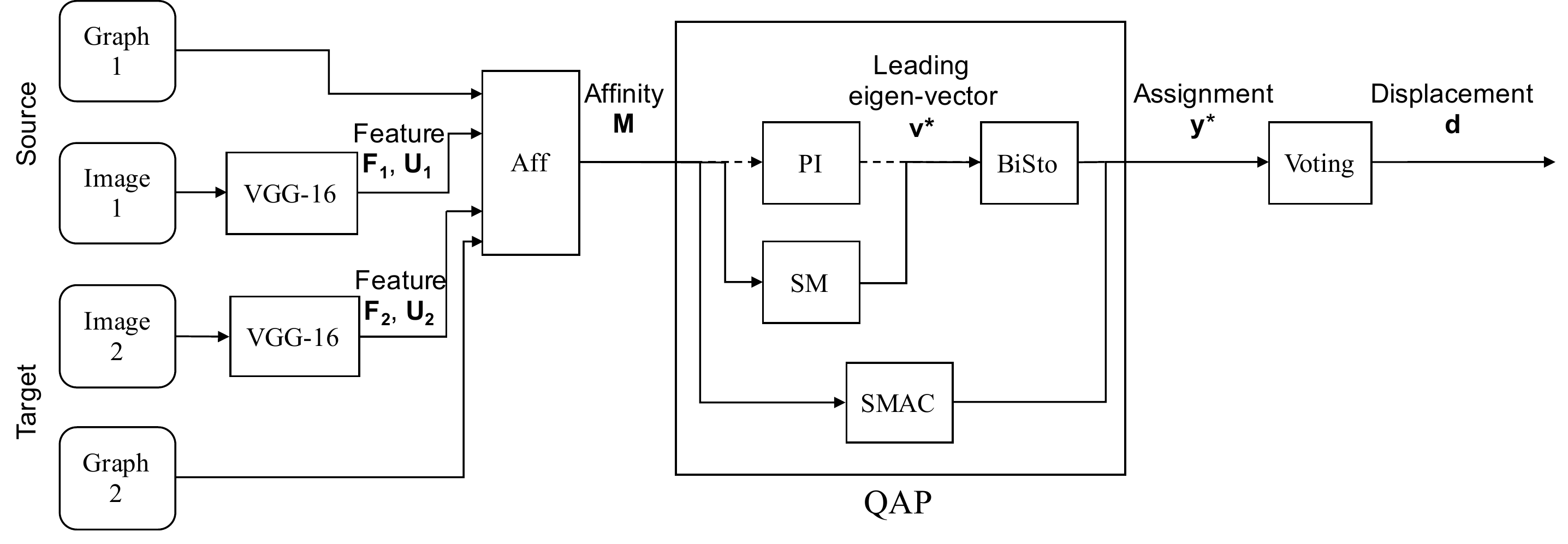}
	\end{subfigure}%
	\caption{Graph matching network. The VGG-16 extracts high level landmark features ($F1, F2, U1, U2$) to compute the affinity matrix $M$.  The upper, middle, and bottom branchs represent solving some relaxed QAP problem with the PI method, the SM layer, and the SMAC layer, respectively.	The Bi-stochastic layer produces a doubly-stochastic confidence map ($C$ in~\eqref{eq:QAP_SMAC}). The voting layer converts a confidence map to predicted placement $d$ (refer to~\cite{zanfir2018deep} for more details).}
	\label{fig:qap_pipeline}
\end{figure*}

We build on the network designed by~\cite{zanfir2018deep}. As shown in figure~\ref{fig:qap_pipeline}, the pipeline takes input a pair of images and for each image it constructs a graph (by e.g., Delaunay triangulation, or fully connecting the nodes) with the landmarks as the nodes of the graph.  At the same time, a CNN backbone (e.g., VGG-16~\cite{simonyan2014very}) extracts for the landmarks high-level features, denoted as $F$ and $U$ in figure~\ref{fig:qap_pipeline}. $F$ and $U$ are respectively used to compute node-to-node affinities and edge-to-edge affinities, and subsequently construct an affinity matrix $M$.  At this point the pipeline diverges into three branches.  The upper dashed line branch represents the original pipeline in~\cite{zanfir2018deep}, in which the QCQP problem~\eqref{eq:rQAP_SM} is solved by the PI algorithm.  The branch in the middle represents using an SM layer to solve~\eqref{eq:rQAP_SM}. The bottom branch represents using an SMAC layer to solve~\eqref{eq:rQAP_SM}.  Note that the output of the SMAC layer is directly doubly-stochastic assignment vector.

\begin{figure}[htbp]
	\centering
	\begin{subfigure}{0.40\textwidth}
		\includegraphics[width=\linewidth]{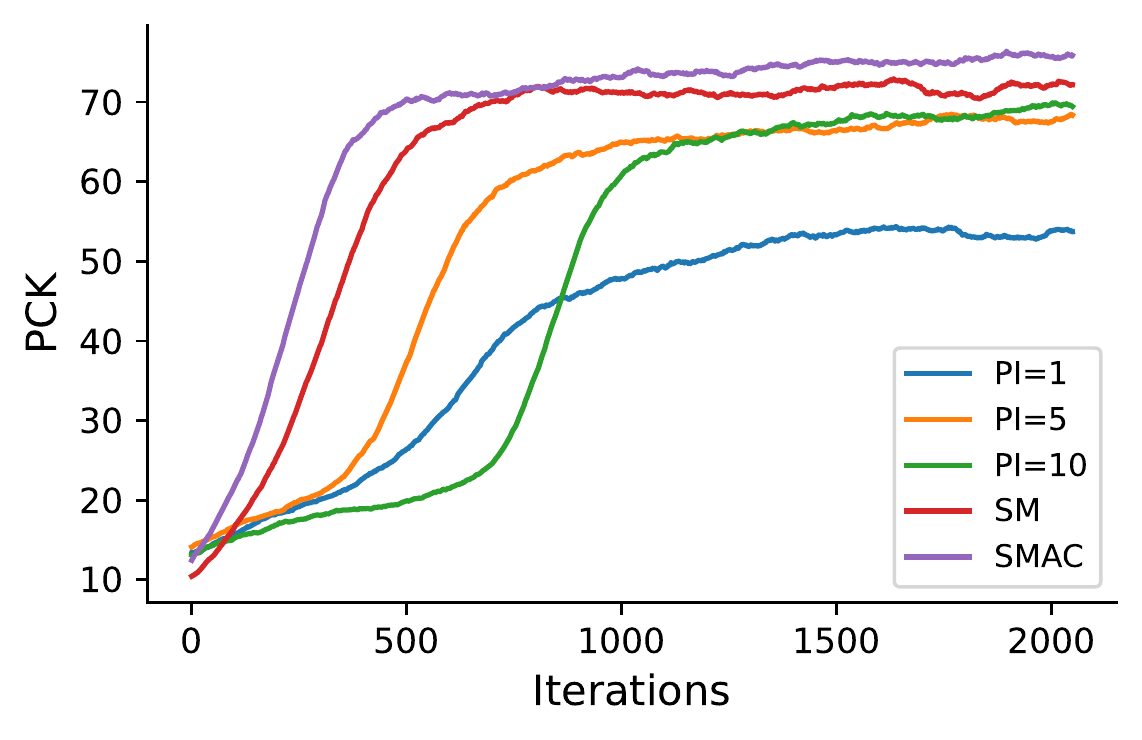}
	\end{subfigure}
	\caption{PCK against training iterations.  $PI=k$ represents the PI method with $k$ iterations.}
	\label{fig:cub_loss_acc}
\end{figure}

The learning curves are plotted in figure~\ref{fig:cub_loss_acc}.  The accuracy metric is defined as the Percentage of Correct Keypoints (PCK)~\cite{yang2011articulated}, by which a match is considered correct if the predicted location is within $\alpha\sqrt{w^2+h^2}$ from the ground truth ($w$ and $h$ are the width and height of the image, respectively, and we set $\alpha=0.1$ throughout the experiment).  It is observed that the implicit layers converge at higher rates than the PI method. 
% (PI=1,5,10), though the eventual accuracy achieved by implicit layers are equivalent to that of PI (again, beating PI is not the purpose of this experiment, and the focus is the versatility of implicit layers).  

Figure~\ref{fig:qap_qualitative} shows qualitative results of the pipeline that uses an SMAC layer. 

%\begin{figure}[ht]
%	\centering
%	\begin{subfigure}{0.40\textwidth}
%		%   \includegraphics[width=\linewidth]{figs/GMN_train_pck_rmean_lr_1e-3.pdf}
%		\includegraphics[width=\linewidth]{figs/GMN_train_pck_rmean_lr_1e-5.pdf}
%	\end{subfigure}
%	\caption{PCK against training iterations.  $PI=k$ represents the PI method with $k$ iterations.}
%	\label{fig:cub_loss_acc}
%\end{figure}

%-------------------------------------------------------------------------

\section{Conclusion}
In this paper we have presented a general treatment of implicitly defined layers 
in feedforward neural networks. The proposed framework, which fits in seamlessly 
with existing explicit formulations, provides a provably richer class of end-to-end trainable neural network architectures. 
We also showed how this framework can be directly incorporated into  
current automatic differentiation techniques for use in backpropagation based 
training. This feature greatly improves the ease-of-use of implicit layers 
by eliminating the need for any additional, problem-specific manual implementation of the backward pass. 
The generality and applicability of implicitly defined layers was demonstrated 
on a number of diverse example problems with very convincing results.

%--------------------------------------------------------------------------------------------
%\begin{acknowledgements}
%\end{acknowledgements}


\bibliographystyle{spmpsci}      % mathematics and physical sciences

\bibliography{implicit_layers, iccv_references}

\begin{thebibliography}{10}
\providecommand{\url}[1]{{#1}}
\providecommand{\urlprefix}{URL }
\expandafter\ifx\csname urlstyle\endcsname\relax
  \providecommand{\doi}[1]{DOI~\discretionary{}{}{}#1}\else
  \providecommand{\doi}{DOI~\discretionary{}{}{}\begingroup
  \urlstyle{rm}\Url}\fi

\bibitem{amos2017optnet}
Amos, B., Kolter, J.Z.: Optnet: Differentiable optimization as a layer in
  neural networks.
\newblock In: Proceedings of the 34th International Conference on Machine
  Learning-Volume 70, pp. 136--145. JMLR. org (2017)

\bibitem{amos2017input}
Amos, B., Xu, L., Kolter, J.Z.: Input convex neural networks.
\newblock In: Proceedings of the 34th International Conference on Machine
  Learning-Volume 70, pp. 146--155. JMLR. org (2017)

\bibitem{bazaraa1982use}
Bazaraa, M.S., Sherali, H.D.: On the use of exact and heuristic cutting plane
  methods for the quadratic assignment problem.
\newblock Journal of the Operational Research Society \textbf{33}(11),
  991--1003 (1982)

\bibitem{bonnans2013perturbation}
Bonnans, J.F., Shapiro, A.: Perturbation analysis of optimization problems.
\newblock Springer Science \& Business Media (2013)

\bibitem{chang2015shapenet}
Chang, A.X., Funkhouser, T., Guibas, L., Hanrahan, P., Huang, Q., Li, Z.,
  Savarese, S., Savva, M., Song, S., Su, H., et~al.: Shapenet: An
  information-rich 3d model repository.
\newblock arXiv preprint arXiv:1512.03012  (2015)

\bibitem{choy20163d}
Choy, C.B., Xu, D., Gwak, J., Chen, K., Savarese, S.: 3d-r2n2: A unified
  approach for single and multi-view 3d object reconstruction.
\newblock In: European conference on computer vision, pp. 628--644. Springer
  (2016)

\bibitem{cour2007balanced}
Cour, T., Srinivasan, P., Shi, J.: Balanced graph matching.
\newblock In: Advances in Neural Information Processing Systems, pp. 313--320
  (2007)

\bibitem{dervieux1980finite}
Dervieux, A., Thomasset, F.: A finite element method for the simulation of a
  rayleigh-taylor instability.
\newblock In: Approximation methods for Navier-Stokes problems, pp. 145--158.
  Springer (1980)

\bibitem{fan2017point}
Fan, H., Su, H., Guibas, L.J.: A point set generation network for 3d object
  reconstruction from a single image.
\newblock In: Proceedings of the IEEE Conference on Computer Vision and Pattern
  Recognition., vol.~2, p.~6 (2017)

\bibitem{girdhar2016learning}
Girdhar, R., Fouhey, D.F., Rodriguez, M., Gupta, A.: Learning a predictable and
  generative vector representation for objects.
\newblock In: European Conference on Computer Vision, pp. 484--499. Springer
  (2016)

\bibitem{gould2016differentiating}
Gould, S., Fernando, B., Cherian, A., Anderson, P., Cruz, R.S., Guo, E.: On
  differentiating parameterized argmin and argmax problems with application to
  bi-level optimization.
\newblock arXiv preprint arXiv:1607.05447  (2016)

\bibitem{hamilton_1983}
Hamilton, A.G.: Numbers, Sets and Axioms: The Apparatus of Mathematics.
\newblock Cambridge University Press (1983)

\bibitem{hansen2011visualization}
Hansen, C.D., Johnson, C.R.: Visualization handbook.
\newblock Elsevier (2011)

\bibitem{johnson2016composing}
Johnson, M.J., Duvenaud, D.K., Wiltschko, A., Adams, R.P., Datta, S.R.:
  Composing graphical models with neural networks for structured
  representations and fast inference.
\newblock In: Advances in neural information processing systems, pp. 2946--2954
  (2016)

\bibitem{kanazawa2016warpnet}
Kanazawa, A., Jacobs, D.W., Chandraker, M.: Warpnet: Weakly supervised matching
  for single-view reconstruction.
\newblock In: Proceedings of the IEEE Conference on Computer Vision and Pattern
  Recognition, pp. 3253--3261 (2016)

\bibitem{krantz2012implicit}
Krantz, S.G., Parks, H.R.: The implicit function theorem: history, theory, and
  applications.
\newblock Springer Science \& Business Media (2012)

\bibitem{kunisch2013bilevel}
Kunisch, K., Pock, T.: A bilevel optimization approach for parameter learning
  in variational models.
\newblock SIAM Journal on Imaging Sciences \textbf{6}(2), 938--983 (2013)

\bibitem{liao2018deep}
Liao, Y., Donn{\'e}, S., Geiger, A.: Deep marching cubes: Learning explicit
  surface representations.
\newblock In: Proceedings of the IEEE Conference on Computer Vision and Pattern
  Recognition, pp. 2916--2925 (2018)

\bibitem{lorensen1987marching}
Lorensen, W.E., Cline, H.E.: Marching cubes: A high resolution 3d surface
  construction algorithm.
\newblock In: ACM siggraph computer graphics, vol.~21, pp. 163--169. ACM (1987)

\bibitem{mairal2011task}
Mairal, J., Bach, F., Ponce, J.: Task-driven dictionary learning.
\newblock IEEE transactions on pattern analysis and machine intelligence
  \textbf{34}(4), 791--804 (2011)

\bibitem{malik2001contourand}
Malik, J., Belongie, S., Leung, T., Shi, J.: Contour and texture analysis for
  image segmentation.
\newblock Int. J. Comput. Vision \textbf{43}(1), 7–27 (2001).
\newblock \doi{10.1023/A:1011174803800}.
\newblock \urlprefix\url{https://doi.org/10.1023/A:1011174803800}

\bibitem{Michalkiewicz_2019_ICCV}
Michalkiewicz, M., Pontes, J.K., Jack, D., Baktashmotlagh, M., Eriksson, A.:
  Implicit surface representations as layers in neural networks.
\newblock In: The IEEE International Conference on Computer Vision (ICCV)
  (2019)

\bibitem{osher1988fronts}
Osher, S., Sethian, J.A.: Fronts propagating with curvature-dependent speed:
  algorithms based on hamilton-jacobi formulations.
\newblock Journal of computational physics \textbf{79}(1), 12--49 (1988)

\bibitem{Park_2019_CVPR}
Park, J.J., Florence, P., Straub, J., Newcombe, R., Lovegrove, S.: Deepsdf:
  Learning continuous signed distance functions for shape representation.
\newblock In: The IEEE Conference on Computer Vision and Pattern Recognition
  (CVPR) (2019)

\bibitem{qi2017pointnet}
Qi, C.R., Su, H., Mo, K., Guibas, L.J.: Pointnet: Deep learning on point sets
  for 3d classification and segmentation.
\newblock Proceedings of the IEEE Conference on Computer Vision and Pattern
  Recognition. \textbf{1}(2), 4 (2017)

\bibitem{rezende2016unsupervised}
Rezende, D.J., Eslami, S.A., Mohamed, S., Battaglia, P., Jaderberg, M., Heess,
  N.: Unsupervised learning of 3d structure from images.
\newblock In: Advances in Neural Information Processing Systems, pp. 4996--5004
  (2016)

\bibitem{richter2018matryoshka}
Richter, S.R., Roth, S.: Matryoshka networks: Predicting 3d geometry via nested
  shape layers.
\newblock In: Proceedings of the IEEE Conference on Computer Vision and Pattern
  Recognition, pp. 1936--1944 (2018)

\bibitem{shi2000normalized}
Shi, J., Malik, J.: Normalized cuts and image segmentation.
\newblock IEEE Transactions on Pattern Analysis and Machine Intelligence
  (2000)

\bibitem{simonyan2014very}
Simonyan, K., Zisserman, A.: Very deep convolutional networks for large-scale
  image recognition.
\newblock arXiv preprint arXiv:1409.1556  (2014)

\bibitem{Song2018Sky}
Song, Y., Luo, H., Ma, J., Bin, H., Chang, Z.: Sky detection in hazy image.
\newblock Sensors \textbf{18}, 1060 (2018).
\newblock \doi{10.3390/s18041060}

\bibitem{WahCUB_200_2011}
Wah, C., Branson, S., Welinder, P., Perona, P., Belongie, S.: {The Caltech-UCSD
  Birds-200-2011 Dataset}.
\newblock Tech. Rep. CNS-TR-2011-001, California Institute of Technology (2011)

\bibitem{yang2011articulated}
Yang, Y., Ramanan, D.: Articulated pose estimation with flexible
  mixtures-of-parts.
\newblock In: CVPR 2011, pp. 1385--1392. IEEE (2011)

\bibitem{yu2003multiclass}
Yu, S.X., Shi, J.: Multiclass spectral clustering.
\newblock In: Proceedings of the Ninth IEEE International Conference on
  Computer Vision - Volume 2, ICCV ’03, p. 313. IEEE Computer Society, USA
  (2003)

\bibitem{zanfir2018deep}
Zanfir, A., Sminchisescu, C.: Deep learning of graph matching.
\newblock In: Proceedings of the IEEE Conference on Computer Vision and Pattern
  Recognition, pp. 2684--2693 (2018)

\bibitem{zhou2012factorized}
Zhou, F., De~la Torre, F.: Factorized graph matching.
\newblock In: 2012 IEEE Conference on Computer Vision and Pattern Recognition,
  pp. 127--134. IEEE (2012)

\end{thebibliography}

\end{document}